\documentclass[11pt]{article}

\usepackage[preprint]{acl}

\usepackage{times}
\usepackage{latexsym}
\usepackage{amsthm}
\usepackage{amsmath}
\usepackage{amssymb}
\usepackage{multirow}
\usepackage{booktabs}
\usepackage{algorithm}
\usepackage{algorithmic}
\usepackage{subcaption}

\newtheorem{proposition}{Proposition}
\newcommand{\ours}{\textsc{RScale}}
\usepackage[T1]{fontenc}

\usepackage[utf8]{inputenc}

\usepackage{microtype}

\usepackage{inconsolata}

\usepackage{graphicx}

\title{\ours: Routing-Aware Token-Adaptive Output Correction\\
  for Dynamic Expert Skipping in Mixture-of-Experts LLMs}

\author{
 \textbf{Zhenhao Zhu\textsuperscript{1,2}\thanks{Equal contribution.}},
 \textbf{Yue Liu\textsuperscript{2}\footnotemark[1]},
 \textbf{Jiaheng Zhang\textsuperscript{2}},
 \textbf{Bryan Hooi\textsuperscript{2}}
\\
\\
 \textsuperscript{1}Tsinghua University,
 \textsuperscript{2}National University of Singapore
\\
}

\begin{document}
\maketitle
\begin{abstract}
Dynamic expert skipping reduces the inference cost of Mixture-of-Experts (MoE) LLMs by executing only a token-dependent subset of the top-$k$ routed experts. Renormalizing the router weights over the surviving experts keeps the mixture convex, but it changes the statistics of the MoE output: the output energy changes by a factor that depends on how much routing mass each token loses. A recent remedy, layer-wise distribution alignment (LDA), matches the per-dimension mean and variance of the pruned output to those of the full top-$k$ output with a single affine map per layer. We show that this global map only corrects the average change, while the actual change varies from token to token. Building on this observation, we propose \ours{}, which modulates the LDA scale with a per-token scalar computed from the router gates, namely the log-ratio of the second moments of the renormalized and original gates. \ours{} is agnostic to the skipping rule and adds only $O(d)$ element-wise work per token. Across five MoE models, five skipping rules, and nine benchmarks covering math, knowledge, code and instruction following, \ours{} improves the average performance over LDA on every model, by up to 3.73 points on Qwen3.6-35B-A3B. \footnote{\url{https://github.com/zzh-thu-22/RScale}}
\end{abstract}

\section{Introduction}

MoE layers scale the parameter count of LLMs while keeping the per-token compute bounded: a router sends each token to $k$ out of $N$ experts and mixes their outputs with the router weights \citep{shazeer2017outrageouslylargeneuralnetworks,fedus2022switchtransformersscalingtrillion,lepikhin2020gshardscalinggiantmodels}. Recent open models such as GLM-5.3 \citep{glm5team2026glm5vibecodingagentic}, Kimi-K3 \citep{kimiteam2026kimik3openfrontier} and DeepSeek-V4 \citep{deepseekai2026deepseekv4highlyefficientmilliontoken} push this design towards many fine-grained experts \citep{dai2024deepseekmoeultimateexpertspecialization}. At serving time the cost of an MoE layer is dominated by loading and running the activated experts, so the number of experts executed per token or per batch is a direct handle on latency and throughput.

\begin{figure}[!t]
    \centering
    \begin{subfigure}{\linewidth}
        \centering
        \includegraphics[width=\linewidth]{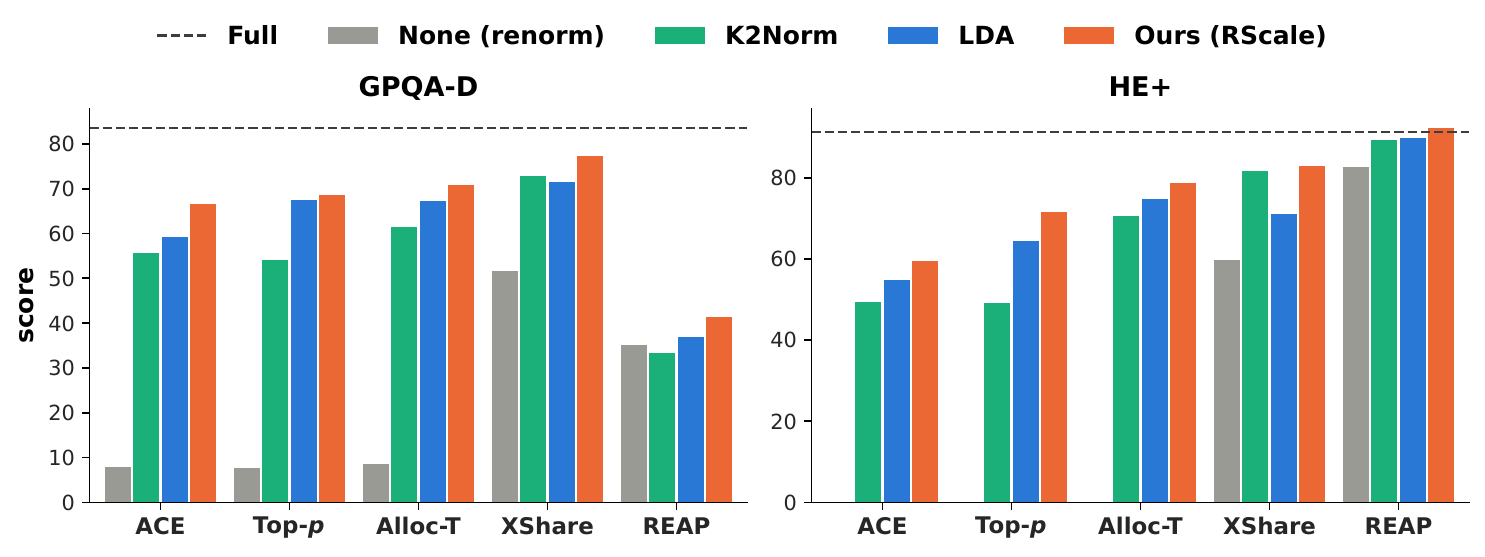}
        \caption{}
        \label{case4a}
    \end{subfigure}

    \vspace{0.5em}

    \begin{subfigure}{\linewidth}
        \centering
        \includegraphics[width=\linewidth]{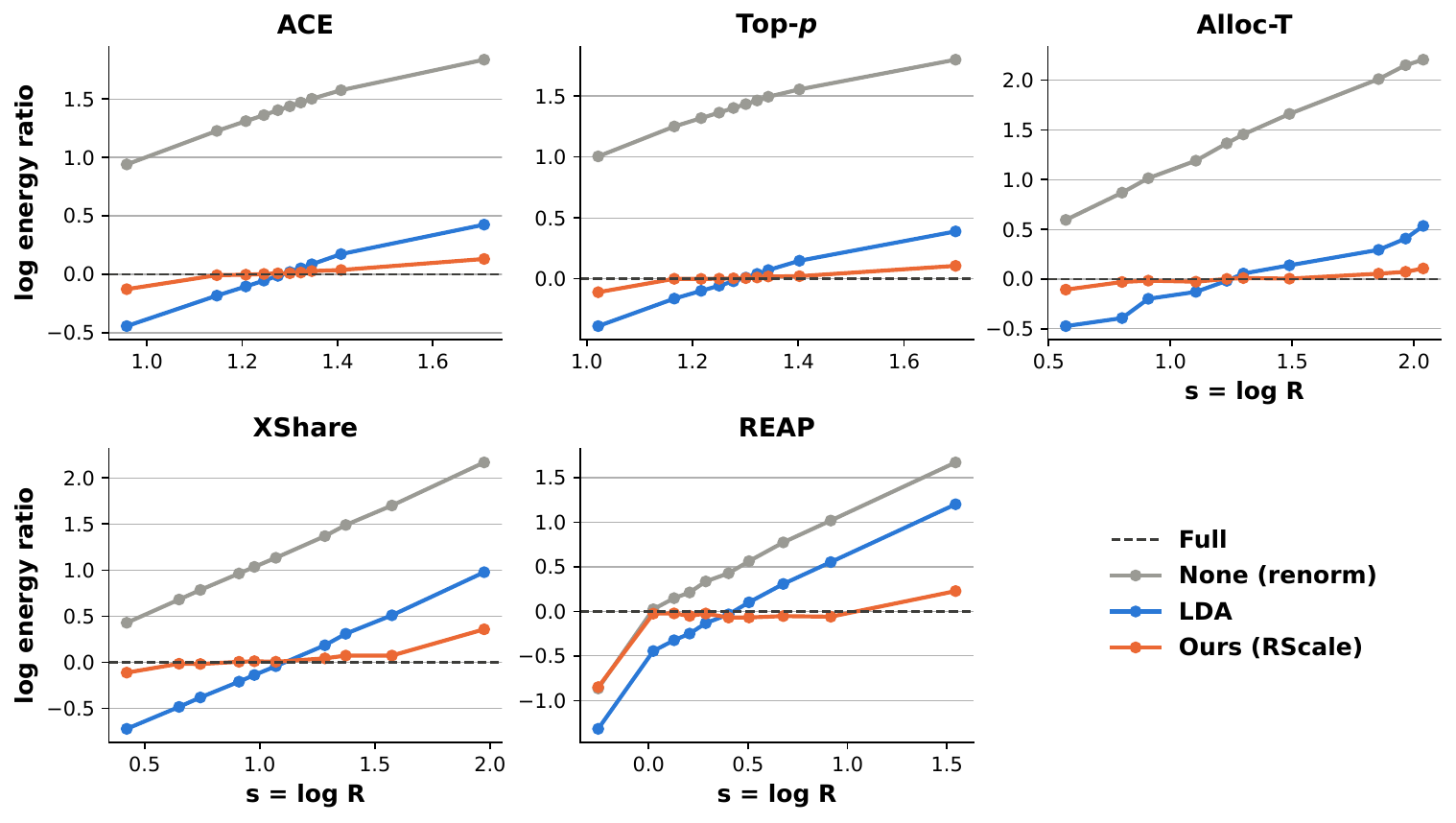}
        \caption{}
        \label{case4b}
    \end{subfigure}
    \caption{Qwen3.6-35B-A3B under five skipping rules. (a) Score with skipping only (None: kept gates renormalized, no output correction) and with K2Norm, LDA and \ours{}; the dashed line is the full model. (b) For each correction, the log-ratio between the energy of $(\mathbf{y}-\boldsymbol{\mu}_f)/\boldsymbol{\sigma}_f$ and that of $(\mathbf{y}_f-\boldsymbol{\mu}_f)/\boldsymbol{\sigma}_f$, where $\mathbf{y}$ is the corrected MoE output. The (token, layer) pairs of all MoE layers are pooled and split into ten equal-count bins of $s$, and both energies are summed within each bin before taking the ratio.  A value of 0 (dashed) means the full energy is restored. Full results for all five models are in Figure~\ref{energy_all}.}
    \label{case4}
\end{figure}

Dynamic expert skipping exploits this handle. Since the routing distribution of a token is often concentrated on a few experts, a large part of the top-$k$ set contributes little to the output \citep{lu2024expertsequalefficientexpert}. Existing methods decide which of the $k$ routed experts to drop with cumulative-probability thresholds \citep{huang2024hardertasksneedexperts}, weight-derived expert contribution scores \citep{xu2026aceadaptivecalibrationfreeexpert}, budgets allocated across the tokens of a batch \citep{liu2026allocmoebudgetawareexpertactivation}, expert sharing within a batch \citep{vankov2026xsharecollaborativeinbatchexpert}, or static expert pruning \citep{lasby2026reap}. Almost all of them renormalize the router weights over the surviving experts so that the mixture stays convex.

Convexity alone, however, does not preserve the output distribution. When a token loses part of its routing mass, the surviving weights are inflated, and the second moment of the MoE output changes by a token-dependent factor. The change accumulates across tens of MoE layers and drives the residual stream away from the regime the model was trained in. The effect is severe in practice: on Qwen3.6-35B-A3B, applying the five skipping rules we study without any output correction lowers the GPQA-Diamond accuracy from 83.58 to 22.16 and the HumanEval+ pass@1 from 91.31 to 28.51 on average (Figure~\ref{case4a}). Two recent works address this gap. \citet{chen2026trainingfreehalvingactivatedexperts} propose to normalize the router weights of the retained experts by the probability mass of a wider top-$k_2$ set, which keeps the gain of the expert branch close to the one seen in training and requires no calibration. \citet{kim2026distributionconsistentinferencedynamicsparse} propose LDA, which estimates the mean and standard deviation of the full and the pruned MoE outputs on calibration data and maps the latter onto the former with a per-layer, per-dimension affine transform.

We find that the latter is more effective. However, it treats the pruned output of a layer as a single population, while the shift that skipping induces is not the same for every token. A token whose routing distribution is peaked loses almost nothing when its tail experts are dropped, whereas a token with a flat distribution may lose most of its mass and receive heavily inflated weights. We formalize this observation in Section~\ref{sec:analysis}. Under the standard approximation of weakly correlated expert outputs, the ratio between the pruned and the full output energy of a token equals $R=\sum_{i\in\mathcal{K}}\tilde{g}_i^2/\sum_{i\in\mathcal{S}}g_i^2$, where $g$ and $\tilde{g}$ are the original and renormalized gates and $\mathcal{S}\supseteq\mathcal{K}$ are the routed and kept sets. A per-layer affine map can only absorb the population average of this ratio, so it under-corrects tokens that lose much mass and over-corrects tokens that lose little. Figure~\ref{case4b} confirms this: after LDA, the output energy is too low for small $s =\log R$ and too high for large $s$.

Building on this analysis, we propose \ours{}. \ours{} keeps the affine form of LDA and multiplies its per-dimension scale by a token-level scalar $c_\ell(s)$, where $s$ is computed from the gate values and the keep mask that the pruned forward pass already produces. For each layer, $c_\ell$ is a piecewise-linear function supported on a few approximately equal-count buckets of $s$. Its knot values are moment-matching estimates computed from the bucket-wise first and second moments of the full and pruned outputs, which we accumulate in a single forward over 8K C4 tokens. A count-based shrinkage towards $c=1$ guards sparsely populated buckets, and $c_\ell\equiv 1$ recovers LDA exactly, so \ours{} is a generalization of LDA.

Our contributions are summarized as follows:
\begin{itemize}
  \item We give an analysis of the distribution shift caused by renormalized expert skipping and identify the per-token energy inflation ratio $R$ as the quantity that a global correction cannot capture.
  \item We propose \ours{}, a training-free, rule-agnostic output correction that modulates the layer-wise distribution alignment with a calibrated function of $\log R$, whose knots are fitted by bucket-wise moment matching in a single calibration pass.
   \item We conduct extensive experiments on five MoE models, five skipping rules, and nine benchmarks covering math, knowledge, code and instruction following. We demonstrate that \ours{} achieves the best average performance among the corrections on every model.
\end{itemize}

\section{Preliminaries and Analysis}
\label{sec:analysis}

\subsection{MoE Layers and Expert Skipping}
\label{sec:prelim}

Consider the MoE layer $\ell$ of a Transformer-based LLM with $N$ routed experts $\{E_i\}_{i=1}^{N}$ and hidden size $d$. For an input token $\mathbf{x}\in\mathbb{R}^d$, the router produces scores $\pi(\mathbf{x})\in\mathbb{R}^N$ and selects the top-$k_0$ set $\mathcal{S}(\mathbf{x})$, $|\mathcal{S}|=k_0$. We write the gates normalized over the selected set as
\begin{equation}
  g_i = \frac{\pi_i}{\sum_{j\in\mathcal{S}}\pi_j}, \qquad i\in\mathcal{S}, \qquad \sum_{i\in\mathcal{S}} g_i = 1.
  \label{eq:gates}
\end{equation}
The routed output and the layer output are
\begin{equation}
  \mathbf{y}_f = \alpha\sum_{i\in\mathcal{S}} g_i\,E_i(\mathbf{x}), \qquad \mathbf{x}' = \mathbf{x} + \mathbf{y}_f + \mathbf{y}_{\mathrm{sh}}(\mathbf{x}),
  \label{eq:moe}
\end{equation}
where $\alpha$ is the routed scaling factor of the model and $\mathbf{y}_{\mathrm{sh}}$ is the output of the shared experts, if any. We call $\mathbf{y}_f$ the \emph{full} output.

\paragraph{Expert skipping as a keep mask.}
A skipping rule assigns each token a keep mask $\kappa\in\{0,1\}^{k_0}$ over the \emph{original} top-$k_0$ slots, with kept set $\mathcal{K}=\{i\in\mathcal{S}:\kappa_i=1\}\neq\emptyset$. Dropped slots are not refilled by lower-ranked experts. The standard practice renormalizes the gates over the survivors,
\begin{equation}
  \tilde{g}_i = \frac{\kappa_i\,g_i}{\sum_{j\in\mathcal{K}} g_j} = \frac{\kappa_i\,g_i}{1-m}, \qquad m = \sum_{j\in\mathcal{S}\setminus\mathcal{K}} g_j,
  \label{eq:renorm}
\end{equation}
where $m\in[0,1)$ is the dropped routing mass, and produces the \emph{pruned} output 
\begin{equation}
  \mathbf{y}_p = \alpha\sum_{i\in\mathcal{K}}\tilde{g}_i\,E_i(\mathbf{x}).
  \label{eq:pruned}
\end{equation}
Throughout the paper, \emph{pruned} refers to quantities computed under dynamic skipping, such as the pruned output, the pruned model and its next-token distribution.

\subsection{Skipping Changes the Output Energy Token by Token}
\label{sec:inflation}

Following LDA, we approximate the expert outputs of a token as having a common squared norm and being mutually orthogonal, $\langle E_i(\mathbf{x}),E_j(\mathbf{x})\rangle\approx \sigma^2(\mathbf{x})\,\delta_{ij}$. Under this approximation, the energy of a mixture with weights $w$ is $\|\alpha\sum_i w_i E_i(\mathbf{x})\|^2\approx\alpha^2 \sigma^2(\mathbf{x})\sum_i w_i^2$, and the energy ratio of the pruned and the full outputs of a token is
\begin{equation}
  \begin{split}
    \frac{\|\mathbf{y}_p\|^2}{\|\mathbf{y}_f\|^2}
    \approx R(\mathbf{x}) &= \frac{\sum_{i\in\mathcal{K}}\tilde{g}_i^2}{\sum_{i\in\mathcal{S}}g_i^2} \\
    &= \frac{1}{(1-m)^2}\cdot\frac{\sum_{i\in\mathcal{K}}g_i^2}{\sum_{i\in\mathcal{S}}g_i^2}.
  \end{split}
  \label{eq:ratio}
\end{equation}
The ratio depends only on the gate values and the keep mask, both of which the pruned forward pass computes anyway. We call a keep mask \emph{order-consistent} if no dropped gate exceeds a kept gate, i.e., $g_j\le g_i$ for all $i\in\mathcal{K}$ and $j\in\mathcal{S}\setminus\mathcal{K}$.

\begin{proposition}
  \label{prop:ratio}
  For every keep mask, $R\in[1/k_0,\,k_0]$. If the keep mask is order-consistent, then $R\ge 1/(1-m)\ge 1$, with $R=1$ if and only if no slot is dropped. 
\end{proposition}

The proof is given in Appendix~\ref{app:proof}. Proposition~\ref{prop:ratio} formalizes the change in output energy that LDA observes at the population level and shows that the size of this change is a property of the individual token. A token with a peaked routing distribution keeps most of its mass and has $R\approx 1$, whereas a token with flat gates that keeps one or two experts can have $R$ close to $k_0$. Rules that are not order-consistent, such as static pruning, may drop the largest gate of a token and also produce $R<1$.

\subsection{Layer-wise Distribution Alignment}
\label{sec:lda}

LDA corrects the pruned output with a per-layer, per-dimension affine map
\begin{equation}
  \mathbf{y}_{\mathrm{lda}} = \boldsymbol{\mu}_f + \mathbf{a}\odot(\mathbf{y}_p-\boldsymbol{\mu}_p), \qquad \mathbf{a} = \frac{\boldsymbol{\sigma}_f}{\boldsymbol{\sigma}_p},
  \label{eq:lda}
\end{equation}

where $(\boldsymbol{\mu}_f,\boldsymbol{\sigma}_f)$ and $(\boldsymbol{\mu}_p,\boldsymbol{\sigma}_p)$ are the per-dimension means and standard deviations of $\mathbf{y}_f$ and $\mathbf{y}_p$ over a calibration set, and all operations are element-wise. By construction, the corrected output has the same first two moments as the full output when averaged over the calibration tokens. However, the map is the same for every token. Ignoring the means, $a_j^2\approx\mathbb{E}_{\mathrm{cal}}[y_{f,j}^2]/\mathbb{E}_{\mathrm{cal}}[y_{p,j}^2]$ for each dimension $j$, where $\mathbb{E}_{\mathrm{cal}}$ denotes the calibration-set average. If we further assume that, given the routing state $(g,\kappa)$, the expert outputs are uncorrelated in each dimension with a constant second moment $\sigma_j^2$, then $\mathbb{E}_{\mathrm{cal}}[y_{f,j}^2]\approx\alpha^2\sigma_j^2\,\mathbb{E}_{\mathrm{cal}}\big[\sum_{i\in\mathcal{S}}g_i^2\big]$ and likewise for $\mathbf{y}_p$. Hence $\sigma_j^2$ cancels and every dimension receives approximately the same squared scale,

\begin{equation}
  a_j^2 \approx \frac{\mathbb{E}_{\mathrm{cal}}\big[\sum_{i\in\mathcal{S}}g_i^2\big]}{\mathbb{E}_{\mathrm{cal}}\big[\sum_{i\in\mathcal{K}}\tilde{g}_i^2\big]} = \frac{1}{\bar{R}},
  \label{eq:lda-scale}
\end{equation}
where $\bar{R}$ is a ratio of calibration averages rather than the average of $R(\mathbf{x})$. For a given token, the corrected output then has $\|\mathbf{y}_{\mathrm{lda}}\|^2\approx\|\mathbf{y}_p\|^2/\bar{R}$, so its residual energy ratio $\|\mathbf{y}_{\mathrm{lda}}\|^2/\|\mathbf{y}_f\|^2$ is about $R(\mathbf{x})/\bar{R}$. Tokens that lose more mass than average remain inflated, and tokens that lose less are deflated. Since $R$ spans the interval $[1,k_0]$ for order-consistent rules, this residual is not small, and a correction that equalizes it must depend on the token. This argument suggests that, under the idealized energy model, restoring the energy of each token requires an additional token-dependent factor on top of the LDA scale, namely $c=(R/\bar{R})^{-1/2}=\exp\big(-(\log R-\log\bar{R})/2\big)$, which is a decreasing function of $\log R$. Real expert outputs are correlated and have non-zero means, so we do not use this closed form. Instead, we estimate the dependence on $\log R$ from data.

\begin{figure*}[t]
    \centering
    \includegraphics[width=\linewidth]{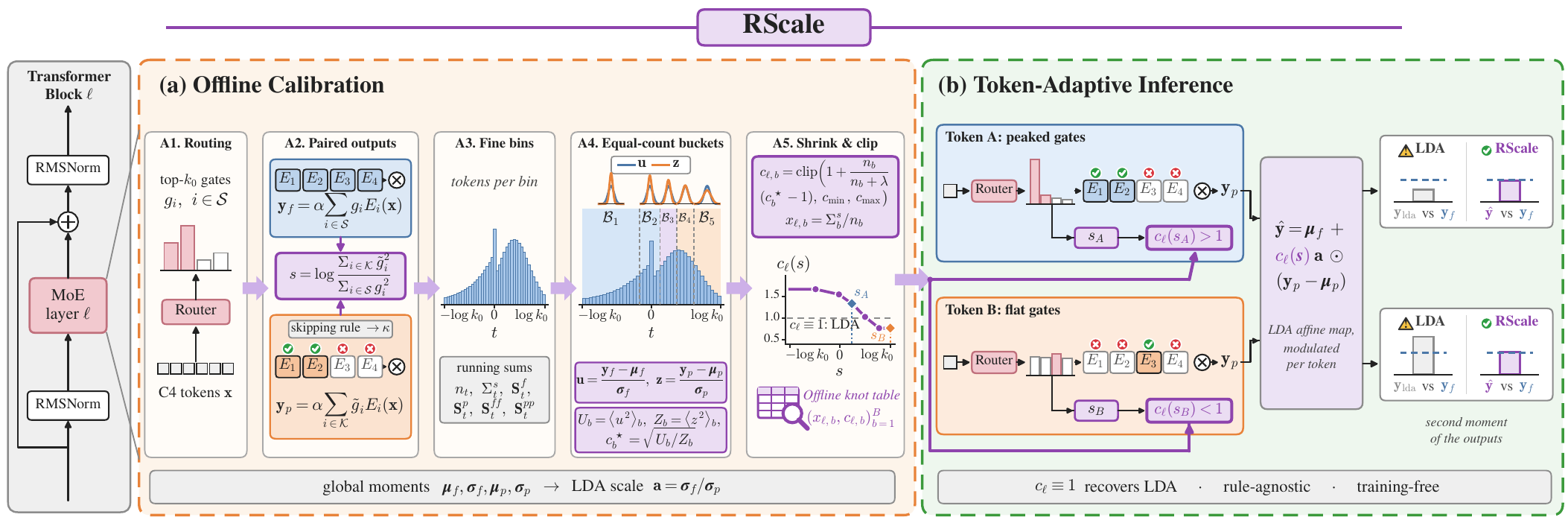}
    \caption{Overview of \ours{} for MoE layer $\ell$. (a) Offline calibration: one forward pass over 8K C4 tokens gives, for each token, the full output $\mathbf{y}_f$, the pruned output $\mathbf{y}_p$ and the routing signal $s=\log R$ (A1--A2). Running sums over fine bins of $t$ (A3) are merged into $B$ approximately equal-count buckets, each with the moment-matching scale $c_b^\star=\sqrt{U_b/Z_b}$ (A4), which is shrunk towards 1 and clipped to give the knots $(x_{\ell,b},c_{\ell,b})$ (A5). (b) Token-adaptive inference: each token computes $s$ from its gates and keep mask, looks up $c_\ell(s)$ in the knot table and multiplies the LDA scale by it. A token with peaked gates (A) gets $c_\ell>1$ and a token with flat gates (B) gets $c_\ell<1$, so the second moment of the corrected output $\hat{\mathbf{y}}$ matches that of the full output $\mathbf{y}_f$, which LDA ($c_\ell\equiv 1$) does not.}
    \label{fig:overview}
\end{figure*}

\section{Method}
\label{sec:method}

\ours{} keeps the affine form of LDA and modulates the scale with a scalar that depends on the routing state of the token (Figure~\ref{fig:overview}).

\subsection{Token-Adaptive Scale Modulation}
\label{sec:modulation}

For each token we compute the routing signal
\begin{equation}
  s(\mathbf{x}) = \log\frac{\sum_{i\in\mathcal{K}}\tilde{g}_i^2}{\sum_{i\in\mathcal{S}}g_i^2} \in [-\log k_0,\ \log k_0],
  \label{eq:signal}
\end{equation}
the log of the energy ratio in Eq.~(\ref{eq:ratio}). The corrected output of layer $\ell$ is
\begin{equation}
  \hat{\mathbf{y}} = \boldsymbol{\mu}_f + c_\ell\big(s(\mathbf{x})\big)\,\mathbf{a}\odot(\mathbf{y}_p-\boldsymbol{\mu}_p),
  \label{eq:rscale}
\end{equation}
where $c_\ell:\mathbb{R}\to[c_{\min},c_{\max}]$ is a scalar function of the signal. We use the log-ratio rather than $R$ itself because $R$ enters the energy multiplicatively, so equal steps in $s$ correspond to equal relative changes in the output scale. The signal is computed from the original gates and the keep mask in $O(k_0)$ operations per token, and it does not require running any of the dropped experts.

We parameterize $c_\ell$ as a piecewise-linear function through $B$ knots $(x_{\ell,b},c_{\ell,b})_{b=1}^{B}$ with $x_{\ell,1}<\dots<x_{\ell,B}$,
\begin{equation}
  c_\ell(s) = c_{\ell,b-1} + \frac{s-x_{\ell,b-1}}{x_{\ell,b}-x_{\ell,b-1}}\big(c_{\ell,b}-c_{\ell,b-1}\big)
  \label{eq:pwl}
\end{equation}
for $s\in[x_{\ell,b-1},x_{\ell,b}]$, and $c_\ell(s)=c_{\ell,1}$ for $s<x_{\ell,1}$ and $c_\ell(s)=c_{\ell,B}$ for $s>x_{\ell,B}$. The function is not constrained to be monotone, so it can absorb deviations from the idealized model in Section~\ref{sec:lda}. Setting $c_\ell\equiv 1$ recovers LDA (Eq.~\ref{eq:lda}).

\subsection{Calibration by Bucket-wise Moment Matching}
\label{sec:fit}
\label{sec:calib}

\paragraph{Paired outputs.}
For each skipping rule, we run the model on the calibration tokens and record, in every MoE layer and for every token, the full output $\mathbf{y}_f$, the pruned output $\mathbf{y}_p$ and the signal $s$. Following LDA \citep{kim2026distributionconsistentinferencedynamicsparse}, every MoE layer passes the full output $\mathbf{y}_f$ on to the next layer, so that the input of each layer is that of the full model and the errors of the preceding layers do not accumulate in the calibration statistics. The pruned output $\mathbf{y}_p$ reuses the outputs of the kept experts that $\mathbf{y}_f$ already computes, so the pass needs no extra expert computation.

\paragraph{Fine bins.}
We fit the knots on buckets of tokens with similar signal and approximately equal counts. Forming such buckets directly requires sorting the tokens by $s$, so the triple $(s,\mathbf{y}_f,\mathbf{y}_p)$ of every calibration token would have to be kept for every layer and rule until the pass ends, i.e., $O(nd)$ memory per layer and rule for $n$ tokens. To avoid storing these triples, we discretize the signal range $[-\log k_0,\log k_0]$ into $T$ uniform fine bins and keep only running sums. For every fine bin $t$ we accumulate the count $n_t$, the signal sum $\Sigma^s_t$ and the following sums:
\begin{equation}
  \begin{aligned}
    &\mathbf{S}^{f}_t = \textstyle\sum\mathbf{y}_f, \quad \mathbf{S}^{p}_t = \textstyle\sum\mathbf{y}_p,  \\ 
    &\mathbf{S}^{ff}_t = \textstyle\sum\mathbf{y}_f^2, \quad \mathbf{S}^{pp}_t = \textstyle\sum\mathbf{y}_p^2,
  \end{aligned}
  \label{eq:sums}
\end{equation}
where the sums run over the tokens in the bin and squares are element-wise. The memory cost is $O(Td)$ per layer and rule, independent of the number of calibration tokens. The global moments are $\boldsymbol{\mu}_f=\sum_t\mathbf{S}^{f}_t/n$ and $\boldsymbol{\sigma}_f^2=\sum_t\mathbf{S}^{ff}_t/n-\boldsymbol{\mu}_f^2$ with $n=\sum_t n_t$, and analogously for the pruned output. They define the LDA map of Eq.~(\ref{eq:lda}), so LDA uses the same pass and the two methods differ only in $c_\ell$.

\paragraph{From fine bins to buckets.}
After the pass, we scan the non-empty fine bins in increasing order of $s$ and greedily merge consecutive bins into buckets $\{\mathcal{B}_b\}$, closing a bucket once its count reaches $n/B$. This yields at most $B$ buckets of approximately equal count. Fewer buckets arise when a single fine bin holds more than $n/B$ tokens, e.g., the tokens that drop no expert and all have $s=0$. Tokens with equal signal are thus never split across buckets, and each bucket yields one knot. The sums of a bucket are the sums of its bins, i.e., $n_b=\sum_{t\in\mathcal{B}_b}n_t$, $\Sigma^s_b=\sum_{t\in\mathcal{B}_b}\Sigma^s_t$ and $\mathbf{S}^{\star}_b=\sum_{t\in\mathcal{B}_b}\mathbf{S}^{\star}_t$ for $\star\in\{f,p,ff,pp\}$.

\paragraph{Standardized bucket statistics.}
We fit $c_\ell$ in the standardized space of the full output, where every dimension has unit variance and contributes equally. For a calibration token with paired outputs $(\mathbf{y}_f,\mathbf{y}_p)$, define
\begin{equation}
  \mathbf{u} = \frac{\mathbf{y}_f-\boldsymbol{\mu}_f}{\boldsymbol{\sigma}_f},\;
   \mathbf{z} = \frac{\mathbf{y}_{\mathrm{lda}}-\boldsymbol{\mu}_f}{\boldsymbol{\sigma}_f} = \frac{\mathbf{y}_p-\boldsymbol{\mu}_p}{\boldsymbol{\sigma}_p},
  \label{eq:std}
\end{equation}
so that the corrected output $\hat{\mathbf{y}}$ of Eq.~(\ref{eq:rscale}), standardized in the same way, $\hat{\mathbf{u}}=(\hat{\mathbf{y}}-\boldsymbol{\mu}_f)/\boldsymbol{\sigma}_f$, equals $c\,\mathbf{z}$. Write $\langle\cdot\rangle_b$ for the average over the tokens in $\mathcal{B}_b$ and over the $d$ dimensions. We use two bucket statistics,
\begin{equation}
  U_b = \langle u^2\rangle_b, \qquad Z_b = \langle z^2\rangle_b.
  \label{eq:bucket-stats}
\end{equation}
Expanding the squares, $U_b$ and $Z_b$ follow from these sums as
\begin{equation}
  \begin{aligned}
    U_b &= \Big\langle\frac{\mathbf{S}^{ff}_b-2\boldsymbol{\mu}_f\mathbf{S}^{f}_b+n_b\boldsymbol{\mu}_f^2}{n_b\,\boldsymbol{\sigma}_f^2}\Big\rangle_d, \\
    Z_b &= \Big\langle\frac{\mathbf{S}^{pp}_b-2\boldsymbol{\mu}_p\mathbf{S}^{p}_b+n_b\boldsymbol{\mu}_p^2}{n_b\,\boldsymbol{\sigma}_p^2}\Big\rangle_d. \\
  \end{aligned}
  \label{eq:bucket-from-sums}
\end{equation}
where products and quotients of vectors are element-wise and $\langle\cdot\rangle_d$ averages over dimensions; see Appendix~\ref{app:sums} for the derivation.

\paragraph{Distribution-consistent scale.}
In the spirit of LDA, we choose the scale that matches the second moment of the corrected output to that of the full output inside each bucket. Since $\langle(c_bz)^2\rangle_b=c_b^2\langle z^2\rangle_b=c_b^2Z_b$ and $\langle u^2\rangle_b=U_b$,
\begin{equation}
  \begin{split}
    \langle(c_bz)^2\rangle_b = \langle u^2\rangle_b
    &\iff c_b^2Z_b = U_b \\
    &\implies c_b^\star = \sqrt{U_b/Z_b}.
  \end{split}
  \label{eq:cstar}
\end{equation}
With $c_b\equiv 1$, which is LDA, the constraint holds on the calibration set as a whole, since $\mathbf{u}$ and $\mathbf{z}$ both have unit second moment there by construction; it is in general violated within individual buckets, and Eq.~(\ref{eq:cstar}) enforces it on every bucket.

\paragraph{Shrinkage and clipping.}
Buckets at the extremes of the signal can contain few tokens, which makes $c_b^\star$ noisy. We shrink each estimate towards the LDA value $c=1$ with a weight that grows with the bucket size $n_b$, clip the result, and place the knot at the mean signal of the bucket:
\begin{equation}
  \begin{split}
    c_{\ell,b} &= \operatorname{clip}\Big(1+\frac{n_b}{n_b+\lambda}\big(c_b^\star-1\big),\ c_{\min},\ c_{\max}\Big), \\
    x_{\ell,b} &= \Sigma^s_b/n_b.
  \end{split}
  \label{eq:shrink}
\end{equation}
As $\lambda\to\infty$, \ours{} reduces to LDA.

\subsection{Inference}
\label{sec:inference}

\begin{algorithm}[t]
  \caption{Forward pass of MoE layer $\ell$ with \ours{} for one token.}
  \label{alg:inference}
  \begin{algorithmic}[1]
    \REQUIRE token $\mathbf{x}$; knots $(x_{\ell,b},c_{\ell,b})_{b=1}^{B}$;\newline $\boldsymbol{\mu}_f$, $\boldsymbol{\mu}_p$ and $\mathbf{a}=\boldsymbol{\sigma}_f/\boldsymbol{\sigma}_p$
    \STATE Route $\mathbf{x}$ to $\mathcal{S}$ with gates $g$ \hfill Eq.~(\ref{eq:gates})
    \STATE Apply the skipping rule to obtain $\kappa$ and $\mathcal{K}$
    \STATE $\tilde{g}\leftarrow\kappa\odot g/(1-m)$ \hfill 
    \STATE $\mathbf{y}_p\leftarrow\alpha\sum_{i\in\mathcal{K}}\tilde{g}_i\,E_i(\mathbf{x})$ 
    \STATE $s\leftarrow\log\big(\sum_{i\in\mathcal{K}}\tilde{g}_i^2/\sum_{i\in\mathcal{S}}g_i^2\big)$ \hfill 
    \STATE $c\leftarrow c_\ell(s)$ 
    \STATE $\hat{\mathbf{y}}\leftarrow\boldsymbol{\mu}_f+c\,\mathbf{a}\odot(\mathbf{y}_p-\boldsymbol{\mu}_p)$ \hfill Eq.~(\ref{eq:rscale})
    \RETURN $\mathbf{x}+\hat{\mathbf{y}}+\mathbf{y}_{\mathrm{sh}}(\mathbf{x})$
  \end{algorithmic}
\end{algorithm}

Algorithm~\ref{alg:inference} summarizes the forward pass, where Line~6 locates $s$ among the knots by binary search. Lines~5--7 add $O(k_0+\log B+d)$ operations per token, which is negligible next to the $O(d\,d_{\mathrm{ff}})$ cost of evaluating a single expert with intermediate size $d_{\mathrm{ff}}$. Besides the per-dimension vectors of LDA, \ours{} stores only the $2B$ knot coordinates per layer and rule.

\begin{table*}[t]
  \centering
  \resizebox{\textwidth}{!}{%
  \begin{tabular}{llcccccccccc}
    \toprule
    \multirow{2}{*}{\textbf{Model}} & \multirow{2}{*}{\textbf{Correction}} & \multicolumn{3}{c}{\textbf{Math}} & \multicolumn{3}{c}{\textbf{Knowledge \& Reasoning}} & \multicolumn{2}{c}{\textbf{Code}} & \textbf{Inst.} & \multirow{2}{*}{\textbf{Avg.}} \\
    \cmidrule(lr){3-5}\cmidrule(lr){6-8}\cmidrule(lr){9-10}\cmidrule(lr){11-11}
    & & MATH-500 & AIME25 & AIME26 & MMLU-Pro & GPQA-D & SuperGPQA & HumanEval+ & LCB & IFEval & \\
    \midrule
    \multirow{4}{*}{Qwen3.6-35B-A3B}
      & Full & 98.10 & 90.83 & 91.67 & 83.70 & 83.58 & 65.34 & 91.31 & 59.25 & 86.37 & 83.35 \\
      \cmidrule(lr){2-12}
      & K2Norm & 76.60 & 36.00 & 35.17 & 68.77 & 55.46 & 46.03 & 68.00 & 31.53 & 71.47 & 54.34 \\
      & LDA & 89.31 & 62.00 & 67.83 & 71.41 & 60.48 & 49.56 & 70.97 & 30.39 & 74.96 & 64.10 \\
      & Ours & \textbf{91.99} & \textbf{66.50} & \textbf{73.17} & \textbf{73.44} & \textbf{64.91} & \textbf{52.29} & \textbf{76.98} & \textbf{34.39} & \textbf{76.82} & \textbf{67.83} \\
    \midrule
    \multirow{4}{*}{Qwen3.5-122B-A10B-FP8}
      & Full & 98.67 & 88.33 & 90.83 & 85.23 & 86.24 & 67.98 & 90.85 & 61.33 & 90.01 & 84.39 \\
      \cmidrule(lr){2-12}
      & K2Norm & 86.84 & 25.50 & 26.42 & 72.93 & 57.92 & 52.74 & 76.57 & 34.88 & 80.08 & 57.10 \\
      & LDA & 94.09 & 58.00 & 64.75 & 77.16 & 69.70 & 57.29 & 80.24 & 39.60 & 83.82 & 69.40 \\
      & Ours & \textbf{95.20} & \textbf{60.42} & \textbf{66.42} & \textbf{78.18} & \textbf{72.26} & \textbf{58.50} & \textbf{82.26} & \textbf{46.34} & \textbf{84.29} & \textbf{71.54} \\
    \midrule
    \multirow{4}{*}{Qwen3-30B-A3B-2507}
      & Full & 97.38 & 62.92 & 75.00 & 73.34 & 54.99 & 51.33 & 91.69 & 43.20 & 87.06 & 70.77 \\
      \cmidrule(lr){2-12}
      & K2Norm & 80.09 & 18.58 & 16.27 & 54.13 & 37.57 & 33.58 & 54.42 & 15.52 & 71.13 & 42.37 \\
      & LDA & 88.79 & 28.75 & 31.67 & 60.74 & 46.73 & 38.74 & 53.03 & 19.51 & 72.73 & 48.97 \\
      & Ours & \textbf{90.25} & \textbf{32.25} & \textbf{36.41} & \textbf{62.60} & \textbf{48.89} & \textbf{40.72} & \textbf{57.96} & \textbf{22.71} & \textbf{74.23} & \textbf{51.78} \\
    \midrule
    \multirow{4}{*}{GLM-4.7-Flash}
      & Full & 88.30 & 87.50 & 86.25 & 64.55 & 45.64 & 38.94 & 74.77 & 18.71 & 80.96 & 65.07 \\
      \cmidrule(lr){2-12}
      & K2Norm & 43.73 & 18.83 & 16.76 & 27.73 & 22.13 & 15.25 & 38.63 & 8.37 & 43.17 & 26.07 \\
      & LDA & 76.30 & 73.75 & 71.42 & 49.88 & 35.48 & 28.34 & 63.86 & \textbf{17.38} & 66.54 & 53.66 \\
      & Ours & \textbf{79.07} & \textbf{75.92} & \textbf{73.67} & \textbf{52.60} & \textbf{36.28} & \textbf{30.35} & \textbf{65.59} & 17.21 & \textbf{67.58} & \textbf{55.36} \\
    \midrule
    \multirow{4}{*}{ERNIE-4.5-21B-A3B-PT}
      & Full & 81.22 & 19.58 & 15.42 & 62.37 & 75.88 & 39.80 & 84.76 & 20.32 & 76.72 & 52.90 \\
      \cmidrule(lr){2-12}
      & K2Norm & 52.63 & 5.42 & 2.92 & 38.53 & 41.14 & 23.00 & 67.26 & 14.79 & 58.89 & 33.84 \\
      & LDA & 64.07 & 7.33 & 5.50 & 48.86 & 50.10 & 29.14 & 71.33 & 14.20 & 66.42 & 39.66 \\
      & Ours & \textbf{67.19} & \textbf{8.50} & \textbf{6.75} & \textbf{50.66} & \textbf{55.19} & \textbf{30.76} & \textbf{74.97} & \textbf{14.80} & \textbf{68.23} & \textbf{41.89} \\
    \bottomrule
  \end{tabular}}
  \caption{Main results. Full is the original model with all $k_0$ experts. For the corrections, each cell is the score (\%) averaged over five skipping rules, and every rule-benchmark score is the mean over 8 seeds. Avg.\ is the mean over the nine benchmarks. The best correction per model and benchmark is in \textbf{bold}. Appendix~\ref{app:per-rule} reports the results of each rule.}
  \label{tab:main}
\end{table*}

\section{Experiments}
\label{sec:experiments}

\subsection{Experimental Setup}
\label{sec:setup}

\begin{table}[t]
  \centering
  \resizebox{\columnwidth}{!}{%
  \begin{tabular}{lcccccc}
    \toprule
    \multirow{2}{*}{\textbf{Model}} & \textbf{Routed} & \textbf{Shared} & \multirow{2}{*}{\textbf{Top-$k_0$}} & \multirow{2}{*}{\textbf{Sparsity}} & \textbf{Params.} & \textbf{Active} \\
    & \textbf{experts} & \textbf{experts} & & & \textbf{(B)} & \textbf{(B)} \\
    \midrule
    Qwen3.6-35B-A3B & 256 & 1 & 8 & 96.50\% & 34.7 & 3.5 \\
    Qwen3.5-122B-A10B-FP8 & 256 & 1 & 8 & 96.50\% & 122.1 & 9.8 \\
    Qwen3-30B-A3B-Instruct-2507 & 128 & 0 & 8 & 93.75\% & 30.5 & 3.4 \\
    GLM-4.7-Flash & 64 & 1 & 4 & 92.31\% & 29.9 & 3.9 \\
    ERNIE-4.5-21B-A3B-PT & 64 & 2 & 6 & 87.88\% & 21.8 & 3.4 \\
    \bottomrule
  \end{tabular}}
  \caption{MoE models included in our study. Sparsity is the fraction of routed and shared experts that are inactive for a token, $1-(k_0+N_{\mathrm{sh}})/(N+N_{\mathrm{sh}})$ with $N_{\mathrm{sh}}$ shared experts. Parameter counts cover the language model and exclude the vision encoder and the multi-token prediction module.}
  \label{tab:models}
\end{table}

\paragraph{Models.}
We evaluate five open MoE LLMs, including Qwen3.6-35B-A3B \citep{qwen36_35b_a3b}, Qwen3.5-122B-A10B-FP8 \citep{qwen3.5}, Qwen3-30B-A3B-Instruct-2507 \citep{qwen3technicalreport}, GLM-4.7-Flash \citep{5team2025glm45agenticreasoningcoding} and ERNIE-4.5-21B-A3B-PT \citep{ernie2025technicalreport}, summarized in Table~\ref{tab:models}. 

\paragraph{Skipping rules.}
To show that the correction is agnostic to how experts are dropped, we pair every correction method with five skipping rules that cover token-level, batch-level and static decisions: ACE \citep{xu2026aceadaptivecalibrationfreeexpert}, Top-$p$ \citep{huang2024hardertasksneedexperts}, Alloc-T \citep{liu2026allocmoebudgetawareexpertactivation}, XShare \citep{vankov2026xsharecollaborativeinbatchexpert} and REAP \citep{lasby2026reap}. None of the rules refills dropped slots. Appendix \ref{app:rules} describes each rule and its parameter settings.

\paragraph{Corrections.}
We compare three output corrections. K2Norm \citep{chen2026trainingfreehalvingactivatedexperts} normalizes the kept router probabilities by the mass of the top-$2k_0$ experts instead of the kept experts; it needs no calibration. LDA \citep{kim2026distributionconsistentinferencedynamicsparse} applies Eq.~(\ref{eq:lda}). \ours{} applies Eq.~(\ref{eq:rscale}). LDA and \ours{} share the same calibration pass on 4 sequences of 2,048 tokens from the C4 training split \citep{raffel2023exploringlimitstransferlearning}. The exception is REAP: we found that scoring its experts on C4 makes the outputs of the pruned model collapse. Following the official setup of REAP, we use the evol-codealpaca \citep{luo2023wizardcoder}. For \ours{} we use $T=100$ fine bins, $B=8$ buckets, $\lambda=100$, and $[c_{\min},c_{\max}]=[0.5,2]$ for every model and rule. We did not tune these values per model.

\paragraph{Benchmarks and protocol.}
We evaluate on nine benchmarks in four groups: math, MATH-500 \citep{hendrycks2021measuringmathematicalproblemsolving}, AIME25 \citep{aime25} and AIME26 \citep{aime26}; knowledge and reasoning, MMLU-Pro \citep{wang2024mmluprorobustchallengingmultitask}, GPQA-Diamond \citep{rein2023gpqagraduatelevelgoogleproofqa} and SuperGPQA \citep{pteam2025supergpqascalingllmevaluation}; code generation, HumanEval+ \citep{liu2023codegeneratedchatgptreally} and LiveCodeBench \citep{jain2024livecodebenchholisticcontaminationfree}; and instruction following, IFEval \citep{zhou2023instructionfollowingevaluationlargelanguage}. Appendix~\ref{app:eval} gives the details.

\paragraph{Implementation.}
We implement all skipping rules and corrections as a plugin for vLLM \citep{kwon2023efficient}. The plugin wraps the MoE forward pass of each layer into a single custom operator that performs routing, masking, expert computation with the Triton fused-MoE kernel, and the correction. The operator is opaque to \texttt{torch.compile}, so the models run with compilation and CUDA graphs enabled.

\subsection{Main Results}
\label{sec:main-results}

Table~\ref{tab:main} compares the three corrections on the five models, with every cell averaged over the five skipping rules. \ours{} has the best average score on all five models. It improves over LDA by 3.73, 2.14, 2.81, 1.70 and 2.23 points on Qwen3.6-35B-A3B, Qwen3.5-122B-A10B-FP8, Qwen3-30B-A3B-Instruct-2507, GLM-4.7-Flash and ERNIE-4.5-21B-A3B-PT, respectively, and is the best correction in 44 of the 45 model and benchmark pairs. The exception is LiveCodeBench on GLM-4.7-Flash, where \ours{} trails LDA by 0.17 points. A paired $t$-test over the 8 seeds shows that the gain in average score is significant on all five models (Appendix~\ref{app:significance}).

Averaged over the five models, the gain over LDA is largest on code (3.27 points over HumanEval+ and LiveCodeBench) and math (2.68 points over MATH-500, AIME25 and AIME26), followed by knowledge and reasoning (2.27 points), and smallest on IFEval (1.34 points). The largest single gains are on AIME and code. On Qwen3.6-35B-A3B, \ours{} improves AIME26 and HumanEval+ by 5.34 and 6.01 points, and on Qwen3.5-122B-A10B-FP8 it improves LiveCodeBench by 6.74 points.

K2Norm has the lowest average score on every model, 5.82 to 27.59 points below LDA and 8.05 to 29.29 points below \ours{}. It falls furthest behind on AIME, e.g., 25.50 vs.\ 60.42 for \ours{} on AIME25 with Qwen3.5-122B-A10B-FP8, and on GLM-4.7-Flash, where it retains 40\% of the full-model average while LDA and \ours{} retain 82\% and 85\%. Changing only the renormalization of the kept gates is therefore not enough, and matching the output statistics on calibration data accounts for most of the recovery.

\begin{figure}[t]
    \centering
    \includegraphics[width=\linewidth]{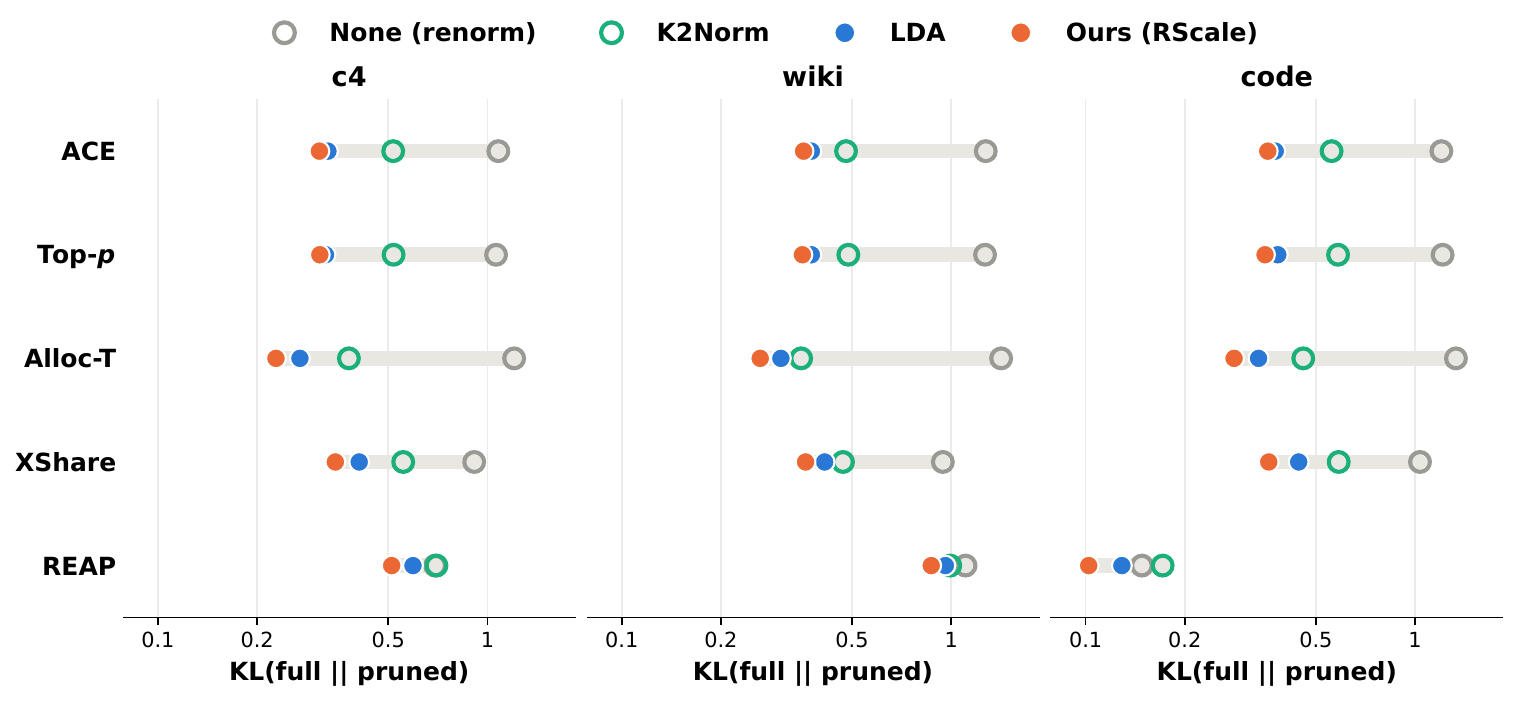}
    \caption{KL divergence from the full to the pruned next-token distribution (log scale, lower is better) on Qwen3.6-35B-A3B, for each skipping rule (rows), held-out corpus (columns) and correction (markers). Full results for all five models are in Figure~\ref{fig:kl-all}.}
    \label{fig:kl}
\end{figure}

\begin{figure}[t]
    \centering
    \includegraphics[width=\linewidth]{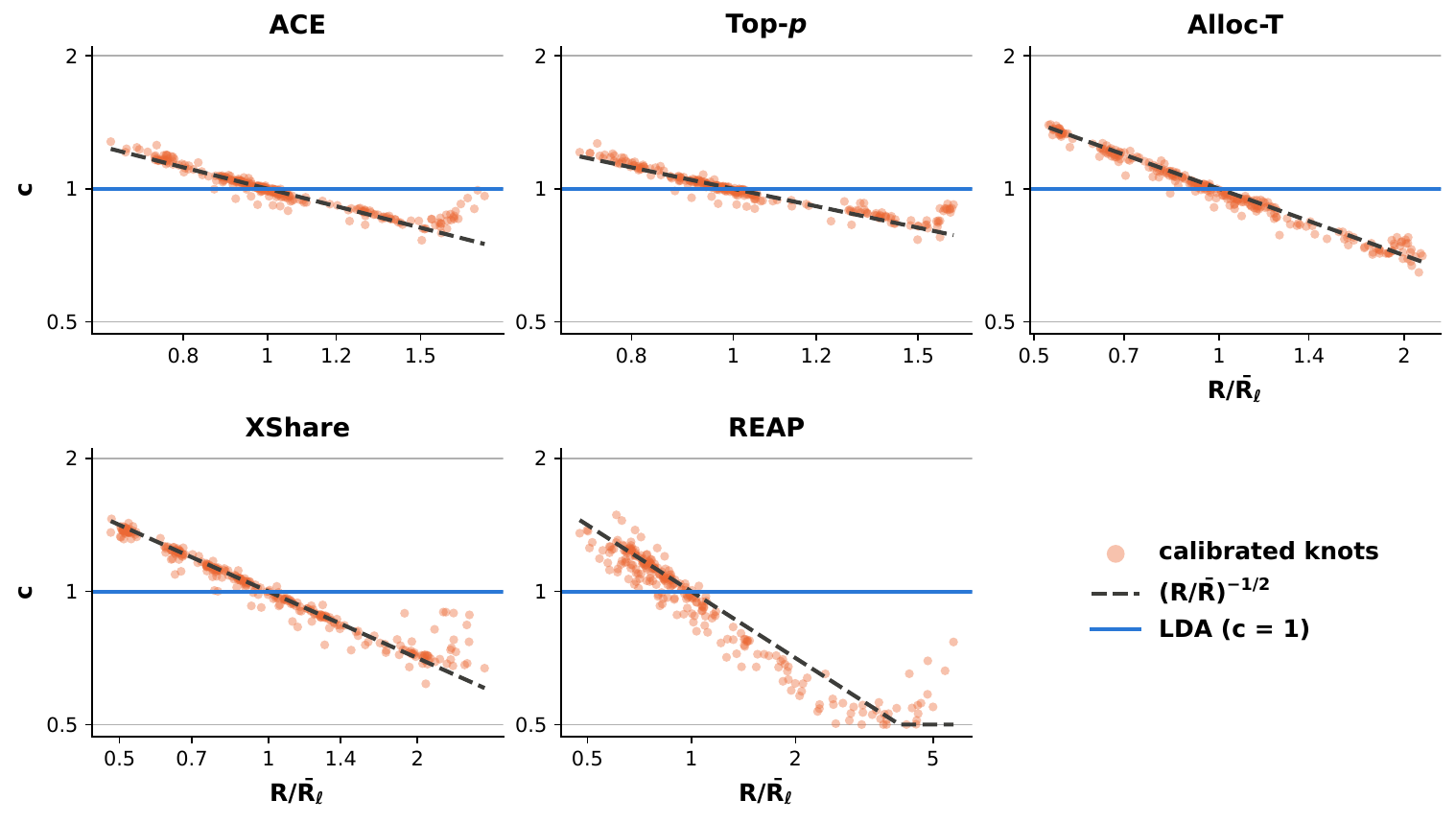}
    \caption{Calibrated knots $(x_{\ell,b},c_{\ell,b})$ of \ours{} on Qwen3.6-35B-A3B, pooled over all MoE layers, plotted against $R/\bar{R}_\ell=\exp(x_{\ell,b})/\bar{R}_\ell$ on log-log axes. The dashed line is the closed form $c=(R/\bar{R}_\ell)^{-1/2}$ of Section~\ref{sec:lda}; LDA corresponds to $c=1$. Full results for all five models are in Figure~\ref{fig:c-all}.}
    \label{fig:c}
\end{figure}

\subsection{Analysis}
\label{sec:fidelity}

\paragraph{Output fidelity.}
To complement the benchmark scores with a measure that does not depend on generation, we also compute how closely the pruned model reproduces the next-token distribution of the full model. For every token we compute the KL divergence $\mathrm{KL}(P_{\mathrm{full}}\,\|\,P_{\mathrm{pruned}})$ over the full vocabulary and average it over the tokens. We use 16 held-out sequences of 2,048 tokens from each of three corpora: C4, disjoint from the calibration sequences; WikiText-2 test \citep{merity2016pointer}, which is not used for calibration; and evol-codealpaca, disjoint from the sequences REAP is calibrated on.

Figure~\ref{fig:kl} shows the results on Qwen3.6-35B-A3B, where \ours{} has the lowest KL in all 15 rule and corpus pairs. Over the five models (Figure~\ref{fig:kl-all}), \ours{} has a lower KL than LDA in 72 of the 75 model, rule and corpus triples, and its KL is on average (geometric mean) 10.7\% lower than that of LDA and 47.7\% lower than that of K2Norm. The reduction over LDA is the same on the three corpora (10.6\%, 10.5\% and 11.1\% on C4, WikiText-2 and code), although the corrections are calibrated on C4 only, except for REAP. It is largest for the batch-level rules XShare (17.5\%) and Alloc-T (16.2\%), whose keep masks spread $R$ more widely than the token-level rules. K2Norm is often worse than no correction at all: its KL is higher than that of None in 12 of the 15 pairs on GLM-4.7-Flash and in 8 of 15 on ERNIE-4.5-21B-A3B-PT. These results agree with the ranking in Table~\ref{tab:main}.

\paragraph{Calibrated scales follow the energy model.}
Section~\ref{sec:lda} predicts that, under the idealized energy model, restoring per-token energy on top of LDA requires the factor $c=(R/\bar{R}_\ell)^{-1/2}$, where $\bar{R}_\ell$ is the ratio of calibration averages in Eq.~(\ref{eq:lda-scale}). Because \ours{} fits its knots from calibration data without using this closed form, we compare the calibrated knots against this reference curve to test whether the empirical scaling follows the predicted power law. We measure $\bar{R}_\ell$ on the calibration tokens of each layer and rule, and plot each knot $c_{\ell,b}$ against $R/\bar{R}_\ell=\exp(x_{\ell,b})/\bar{R}_\ell$. Figure~\ref{fig:c} shows the knots of all MoE layers of Qwen3.6-35B-A3B. Across layers they collapse onto a single decreasing curve close to the prediction, whereas LDA fixes $c=1$ and misses this dependence entirely. To quantify the agreement, we fit $\log c_{\ell,b}$ against $\log(R/\bar{R}_\ell)$ by least squares, weighting each knot by its bucket size $n_b$. The fitted slope is between $-0.55$ and $-0.45$ for 24 of the 25 model and rule pairs, with a median of $-0.501$; the only exception is ACE on Qwen3.5-122B-A10B-FP8, with a slope of $-0.628$. Departures from the reference curve appear mainly in the high-$R/\bar{R}_\ell$ tail for two expected reasons: sparsely populated tail buckets (e.g., at the largest $R/\bar{R}_\ell$ under ACE and XShare) are pulled towards $c=1$ by the shrinkage in Eq.~(\ref{eq:shrink}), producing a mild upturn, while extreme buckets under REAP hit the clipping bound $c_{\min}=0.5$. Beyond these boundary effects, fitting the knots from data also absorbs departures from the idealized assumptions, which is why the calibrated knots outperform applying the closed form directly.

\paragraph{Ablations, budgets and inference cost.}
Appendix~\ref{sec:ablation} ablates the form of the scale function, the shrinkage strength, the number of buckets, and the size and domain of the calibration data. Appendix~\ref{app:budget} shows that \ours{} improves on LDA at every skipping budget, with larger gains for more aggressive budgets. Appendix~\ref{app:efficiency} shows that \ours{} adds negligible serving overhead over LDA and keeps most of the speedup of skipping.

\section{Related Work}
\label{sec:related}

\paragraph{Sparsely activated MoE architecture.}
Mixture-of-experts models combine expert networks through a learned gating function \citep{10.1162/neco.1994.6.2.181,10.1162/neco.1991.3.1.79}. Sparse MoE LLMs activate only a few experts per token, and GLaM \citep{du2022glamefficientscalinglanguage} outperforms the dense GPT-3 with about one third of its training energy. Prior work has studied alternative routing schemes \citep{lewis2021baselayerssimplifyingtraining,roller2021hashlayerslargesparse,zhou2022mixtureofexpertsexpertchoicerouting} and training stability \citep{zoph2022stmoedesigningstabletransferable}, and recent open models scale this design further \citep{deepseekai2024deepseekv2strongeconomicalefficient,deepseekai2025deepseekv3technicalreport}, with DeepSeek-V3 balancing the expert load mainly through per-expert biases on the routing scores \citep{wang2024auxiliarylossfreeloadbalancingstrategy}. The models we study use token-choice top-$k$ routing, so every token activates the same number $k_0$ of routed experts.

\paragraph{Static expert pruning and merging.}
A first line of work permanently removes or merges experts. EAN \citep{jaiswal2025findingfantasticexpertsmoes} compares a range of expert dropping criteria and finds that the expert activation norm is a strong criterion. REAP \citep{lasby2026reap} scores experts by their router-weighted activation norm on domain calibration data and argues that pruning preserves the router's control better than merging. Merging approaches such as M-SMoE \citep{li2024mergecompressdemystifyefficient} and HC-SMoE \citep{chen2025retrainingfreemergingsparsemoe} group similar experts and fuse their weights. These methods reduce memory, but a static pruned set cannot adapt to individual tokens.

\paragraph{Dynamic expert skipping.}
A second line of work keeps all experts and lets each token execute fewer of them. Top-$p$ routing \citep{huang2024hardertasksneedexperts} keeps the smallest prefix of the sorted routing distribution whose cumulative probability exceeds a threshold, and AdaMoE \citep{zeng2024adamoetokenadaptiveroutingnull} learns null experts so that tokens may use fewer real ones; both require training. Among training-free methods, ACE \citep{xu2026aceadaptivecalibrationfreeexpert} augments the router weights with weight-only estimates of the expert output amplitude and skips the slots whose combined importance falls below a cumulative threshold. Alloc-MoE \citep{liu2026allocmoebudgetawareexpertactivation} distributes a global activation budget across the tokens of a batch. At the batch level, XShare \citep{vankov2026xsharecollaborativeinbatchexpert} restrict a batch to a shared expert set chosen from the batch's routing scores, which reduces the number of distinct experts loaded per step.

\section{Conclusion}
\label{sec:conclusion}

We showed that renormalized expert skipping changes the energy of the MoE output by a token-dependent factor $R$, which a per-layer affine correction such as LDA can only match on average. \ours{} modulates the LDA scale with a calibrated function of $\log R$ whose knots are computed directly from bucket-wise moment matching, and reduces to LDA when $c_\ell\equiv 1$. Across five MoE models, five skipping rules and nine benchmarks, it improves the average score over LDA by 1.70 to 3.73 points on every model at negligible cost.

\section*{Limitations}
\ours{} modulates the scale with one scalar per token and does not change the per-dimension shape of the correction, which may explain the cases where it does not improve over LDA. Under the skipping budgets we study, a clear gap to the full model remains (e.g., 67.83 vs.\ 83.35 average score on Qwen3.6-35B-A3B), since \ours{} rescales the pruned output token by token but cannot recover the information carried by the dropped experts.


\bibliography{custom}

@misc{kim2026distributionconsistentinferencedynamicsparse,
      title={Distribution-Consistent Inference for Dynamic Sparse Mixture-of-Experts}, 
      author={Dohyeon Kim and Bedionita Soro and Sung Ju Hwang},
      year={2026},
      eprint={2609.09241},
      archivePrefix={arXiv},
      primaryClass={cs.LG},
      url={https://arxiv.org/abs/2609.09241}, 
}

@misc{5team2025glm45agenticreasoningcoding,
      title={GLM-4.5: Agentic, Reasoning, and Coding (ARC) Foundation Models}, 
      author={GLM Team},
      year={2025},
      eprint={2508.06471},
      archivePrefix={arXiv},
      primaryClass={cs.CL},
      url={https://arxiv.org/abs/2508.06471}, 
}

@misc{ernie2025technicalreport,
      title={ERNIE 4.5 Technical Report},
      author={Baidu ERNIE Team},
      year={2025},
      eprint={},
      archivePrefix={arXiv},
      primaryClass={cs.CL},
      url={}
}

@misc{qwen3technicalreport,
      title={Qwen3 Technical Report}, 
      author={Qwen Team},
      year={2025},
      eprint={2505.09388},
      archivePrefix={arXiv},
      primaryClass={cs.CL},
      url={https://arxiv.org/abs/2505.09388}, 
}

@misc{qwen3.5,
    title  = {{Qwen3.5}: Towards Native Multimodal Agents},
    author = {{Qwen Team}},
    month  = {February},
    year   = {2026},
    url    = {https://qwen.ai/blog?id=qwen3.5}
}

@misc{qwen36_35b_a3b,
    title = {{Qwen3.6-35B-A3B}: Agentic Coding Power, Now Open to All},
    url = {https://qwen.ai/blog?id=qwen3.6-35b-a3b},
    author = {{Qwen Team}},
    month = {April},
    year = {2026}
}

@misc{hendrycks2021measuringmathematicalproblemsolving,
      title={Measuring Mathematical Problem Solving With the MATH Dataset}, 
      author={Dan Hendrycks and Collin Burns and Saurav Kadavath and Akul Arora and Steven Basart and Eric Tang and Dawn Song and Jacob Steinhardt},
      year={2021},
      eprint={2103.03874},
      archivePrefix={arXiv},
      primaryClass={cs.LG},
      url={https://arxiv.org/abs/2103.03874}, 
}

@misc{wang2024mmluprorobustchallengingmultitask,
      title={MMLU-Pro: A More Robust and Challenging Multi-Task Language Understanding Benchmark}, 
      author={Yubo Wang and Xueguang Ma and Ge Zhang and Yuansheng Ni and Abhranil Chandra and Shiguang Guo and Weiming Ren and Aaran Arulraj and Xuan He and Ziyan Jiang and Tianle Li and Max Ku and Kai Wang and Alex Zhuang and Rongqi Fan and Xiang Yue and Wenhu Chen},
      year={2024},
      eprint={2406.01574},
      archivePrefix={arXiv},
      primaryClass={cs.CL},
      url={https://arxiv.org/abs/2406.01574}, 
}

@misc{rein2023gpqagraduatelevelgoogleproofqa,
      title={GPQA: A Graduate-Level Google-Proof Q\&A Benchmark}, 
      author={David Rein and Betty Li Hou and Asa Cooper Stickland and Jackson Petty and Richard Yuanzhe Pang and Julien Dirani and Julian Michael and Samuel R. Bowman},
      year={2023},
      eprint={2311.12022},
      archivePrefix={arXiv},
      primaryClass={cs.AI},
      url={https://arxiv.org/abs/2311.12022}, 
}

@misc{pteam2025supergpqascalingllmevaluation,
      title={SuperGPQA: Scaling LLM Evaluation across 285 Graduate Disciplines}, 
      author={M-A-P Team},
      year={2025},
      eprint={2502.14739},
      archivePrefix={arXiv},
      primaryClass={cs.CL},
      url={https://arxiv.org/abs/2502.14739}, 
}

@misc{liu2023codegeneratedchatgptreally,
      title={Is Your Code Generated by ChatGPT Really Correct? Rigorous Evaluation of Large Language Models for Code Generation}, 
      author={Jiawei Liu and Chunqiu Steven Xia and Yuyao Wang and Lingming Zhang},
      year={2023},
      eprint={2305.01210},
      archivePrefix={arXiv},
      primaryClass={cs.SE},
      url={https://arxiv.org/abs/2305.01210}, 
}

@misc{jain2024livecodebenchholisticcontaminationfree,
      title={LiveCodeBench: Holistic and Contamination Free Evaluation of Large Language Models for Code}, 
      author={Naman Jain and King Han and Alex Gu and Wen-Ding Li and Fanjia Yan and Tianjun Zhang and Sida Wang and Armando Solar-Lezama and Koushik Sen and Ion Stoica},
      year={2024},
      eprint={2403.07974},
      archivePrefix={arXiv},
      primaryClass={cs.SE},
      url={https://arxiv.org/abs/2403.07974}, 
}

@misc{zhou2023instructionfollowingevaluationlargelanguage,
      title={Instruction-Following Evaluation for Large Language Models}, 
      author={Jeffrey Zhou and Tianjian Lu and Swaroop Mishra and Siddhartha Brahma and Sujoy Basu and Yi Luan and Denny Zhou and Le Hou},
      year={2023},
      eprint={2311.07911},
      archivePrefix={arXiv},
      primaryClass={cs.CL},
      url={https://arxiv.org/abs/2311.07911}, 
}

@misc{aime25,
      title={American Invitational Mathematics Examination (AIME) 2025}, 
      author={Zhang, Yifan and Math-AI, Team},
      year={2025},
}

@misc{aime26,
      title={American Invitational Mathematics Examination (AIME) 2026}, 
      author={Zhang, Yifan and Math-AI, Team},
      year={2026},
}

@misc{evalscope_2024,
    title={{EvalScope}: Evaluation Framework for Large Models},
    author={ModelScope Team},
    year={2024},
    url={https://github.com/modelscope/evalscope}
}

@inproceedings{kwon2023efficient,
  title={Efficient Memory Management for Large Language Model Serving with PagedAttention},
  author={Woosuk Kwon and Zhuohan Li and Siyuan Zhuang and Ying Sheng and Lianmin Zheng and Cody Hao Yu and Joseph E. Gonzalez and Hao Zhang and Ion Stoica},
  booktitle={Proceedings of the ACM SIGOPS 29th Symposium on Operating Systems Principles},
  year={2023}
}

@misc{liu2026allocmoebudgetawareexpertactivation,
      title={Alloc-MoE: Budget-Aware Expert Activation Allocation for Efficient Mixture-of-Experts Inference}, 
      author={Baihui Liu and Kaiyuan Tian and Wei Wang and Zhaoning Zhang and Linbo Qiao and Dongsheng Li},
      year={2026},
      eprint={2604.08133},
      archivePrefix={arXiv},
      primaryClass={cs.LG},
      url={https://arxiv.org/abs/2604.08133}, 
}

@misc{lu2024expertsequalefficientexpert,
      title={Not All Experts are Equal: Efficient Expert Pruning and Skipping for Mixture-of-Experts Large Language Models}, 
      author={Xudong Lu and Qi Liu and Yuhui Xu and Aojun Zhou and Siyuan Huang and Bo Zhang and Junchi Yan and Hongsheng Li},
      year={2024},
      eprint={2402.14800},
      archivePrefix={arXiv},
      primaryClass={cs.CL},
      url={https://arxiv.org/abs/2402.14800}, 
}

@inproceedings{
    lasby2026reap,
    title={{REAP} the Experts: Why Pruning Prevails for One-Shot MoE compression},
    author={Mike Lasby and Ivan Lazarevich and Nish Sinnadurai and Sean Lie and Yani Ioannou and Vithursan Thangarasa},
    booktitle={The Fourteenth International Conference on Learning Representations},
    year={2026},
    url={https://openreview.net/forum?id=ukGxWd2aDG}
}

@misc{vankov2026xsharecollaborativeinbatchexpert,
      title={XShare: Collaborative in-Batch Expert Sharing for Faster MoE Inference}, 
      author={Daniil Vankov and Nikita Ivkin and Kyle Ulrich and Xiang Song and Ashish Khetan and George Karypis},
      year={2026},
      eprint={2602.07265},
      archivePrefix={arXiv},
      primaryClass={cs.LG},
      url={https://arxiv.org/abs/2602.07265}, 
}

@misc{xu2026aceadaptivecalibrationfreeexpert,
      title={ACE: Adaptive Calibration-Free Expert Skipping for MoE-based LLMs}, 
      author={Zukang Xu and Zhixiong Zhao and Xing Hu and Jiangyong Yu and Houji Wen and Jun Li and Zhe Jiang and Dawei Yang},
      year={2026},
      eprint={2609.05228},
      archivePrefix={arXiv},
      primaryClass={cs.AI},
      url={https://arxiv.org/abs/2609.05228}, 
}

@misc{huang2024hardertasksneedexperts,
      title={Harder Tasks Need More Experts: Dynamic Routing in MoE Models}, 
      author={Quzhe Huang and Zhenwei An and Nan Zhuang and Mingxu Tao and Chen Zhang and Yang Jin and Kun Xu and Kun Xu and Liwei Chen and Songfang Huang and Yansong Feng},
      year={2024},
      eprint={2403.07652},
      archivePrefix={arXiv},
      primaryClass={cs.LG},
      url={https://arxiv.org/abs/2403.07652}, 
}

@misc{chen2026trainingfreehalvingactivatedexperts,
      title={Training-Free Halving of Activated Experts in Fine-Grained Mixture-of-Experts Models}, 
      author={Xing Chen and Hengshuai Yao},
      year={2026},
      eprint={2609.04575},
      archivePrefix={arXiv},
      primaryClass={cs.LG},
      url={https://arxiv.org/abs/2609.04575}, 
}

@misc{luo2023wizardcoder,
      title={WizardCoder: Empowering Code Large Language Models with Evol-Instruct}, 
      author={Ziyang Luo and Can Xu and Pu Zhao and Qingfeng Sun and Xiubo Geng and Wenxiang Hu and Chongyang Tao and Jing Ma and Qingwei Lin and Daxin Jiang},
      year={2023},
}

@misc{raffel2023exploringlimitstransferlearning,
      title={Exploring the Limits of Transfer Learning with a Unified Text-to-Text Transformer}, 
      author={Colin Raffel and Noam Shazeer and Adam Roberts and Katherine Lee and Sharan Narang and Michael Matena and Yanqi Zhou and Wei Li and Peter J. Liu},
      year={2023},
      eprint={1910.10683},
      archivePrefix={arXiv},
      primaryClass={cs.LG},
      url={https://arxiv.org/abs/1910.10683}, 
}

@misc{li2024mergecompressdemystifyefficient,
      title={Merge, Then Compress: Demystify Efficient SMoE with Hints from Its Routing Policy}, 
      author={Pingzhi Li and Zhenyu Zhang and Prateek Yadav and Yi-Lin Sung and Yu Cheng and Mohit Bansal and Tianlong Chen},
      year={2024},
      eprint={2310.01334},
      archivePrefix={arXiv},
      primaryClass={cs.LG},
      url={https://arxiv.org/abs/2310.01334}, 
}

@misc{chen2025retrainingfreemergingsparsemoe,
      title={Retraining-Free Merging of Sparse MoE via Hierarchical Clustering}, 
      author={I-Chun Chen and Hsu-Shen Liu and Wei-Fang Sun and Chen-Hao Chao and Yen-Chang Hsu and Chun-Yi Lee},
      year={2025},
      eprint={2410.08589},
      archivePrefix={arXiv},
      primaryClass={cs.LG},
      url={https://arxiv.org/abs/2410.08589}, 
}

@misc{jaiswal2025findingfantasticexpertsmoes,
      title={Finding Fantastic Experts in MoEs: A Unified Study for Expert Dropping Strategies and Observations}, 
      author={Ajay Jaiswal and Jianyu Wang and Yixiao Li and Pingzhi Li and Tianlong Chen and Zhangyang Wang and Chong Wang and Ruoming Pang and Xianzhi Du},
      year={2025},
      eprint={2504.05586},
      archivePrefix={arXiv},
      primaryClass={cs.LG},
      url={https://arxiv.org/abs/2504.05586}, 
}

@misc{zeng2024adamoetokenadaptiveroutingnull,
      title={AdaMoE: Token-Adaptive Routing with Null Experts for Mixture-of-Experts Language Models}, 
      author={Zihao Zeng and Yibo Miao and Hongcheng Gao and Hao Zhang and Zhijie Deng},
      year={2024},
      eprint={2406.13233},
      archivePrefix={arXiv},
      primaryClass={cs.AI},
      url={https://arxiv.org/abs/2406.13233}, 
}

@misc{shazeer2017outrageouslylargeneuralnetworks,
      title={Outrageously Large Neural Networks: The Sparsely-Gated Mixture-of-Experts Layer}, 
      author={Noam Shazeer and Azalia Mirhoseini and Krzysztof Maziarz and Andy Davis and Quoc Le and Geoffrey Hinton and Jeff Dean},
      year={2017},
      eprint={1701.06538},
      archivePrefix={arXiv},
      primaryClass={cs.LG},
      url={https://arxiv.org/abs/1701.06538}, 
}

@misc{fedus2022switchtransformersscalingtrillion,
      title={Switch Transformers: Scaling to Trillion Parameter Models with Simple and Efficient Sparsity}, 
      author={William Fedus and Barret Zoph and Noam Shazeer},
      year={2022},
      eprint={2101.03961},
      archivePrefix={arXiv},
      primaryClass={cs.LG},
      url={https://arxiv.org/abs/2101.03961}, 
}

@misc{lepikhin2020gshardscalinggiantmodels,
      title={GShard: Scaling Giant Models with Conditional Computation and Automatic Sharding}, 
      author={Dmitry Lepikhin and HyoukJoong Lee and Yuanzhong Xu and Dehao Chen and Orhan Firat and Yanping Huang and Maxim Krikun and Noam Shazeer and Zhifeng Chen},
      year={2020},
      eprint={2006.16668},
      archivePrefix={arXiv},
      primaryClass={cs.CL},
      url={https://arxiv.org/abs/2006.16668}, 
}

@misc{deepseekai2026deepseekv4highlyefficientmilliontoken,
      title={DeepSeek-V4: Towards Highly Efficient Million-Token Context Intelligence}, 
      author={DeepSeek-AI},
      year={2026},
      eprint={2606.19348},
      archivePrefix={arXiv},
      primaryClass={cs.CL},
      url={https://arxiv.org/abs/2606.19348}, 
}

@misc{kimiteam2026kimik3openfrontier,
      title={Kimi K3: Open Frontier Intelligence}, 
      author={Kimi Team},
      year={2026},
      eprint={2607.24653},
      archivePrefix={arXiv},
      primaryClass={cs.CL},
      url={https://arxiv.org/abs/2607.24653}, 
}

@misc{glm5team2026glm5vibecodingagentic,
      title={GLM-5: from Vibe Coding to Agentic Engineering},
      author={GLM-5-Team},
      year={2026},
      eprint={2602.15763},
      archivePrefix={arXiv},
      primaryClass={cs.LG},
      url={https://arxiv.org/abs/2602.15763},
}

@misc{merity2016pointer,
      title={Pointer Sentinel Mixture Models},
      author={Stephen Merity and Caiming Xiong and James Bradbury and Richard Socher},
      year={2016},
      eprint={1609.07843},
      archivePrefix={arXiv},
      primaryClass={cs.CL}
}

@misc{paster2023openwebmath,
      title={OpenWebMath: An Open Dataset of High-Quality Mathematical Web Text},
      author={Keiran Paster and Marco Dos Santos and Zhangir Azerbayev and Jimmy Ba},
      year={2023},
      eprint={2310.06786},
      archivePrefix={arXiv},
      primaryClass={cs.AI}
}

@misc{dai2024deepseekmoeultimateexpertspecialization,
      title={DeepSeekMoE: Towards Ultimate Expert Specialization in Mixture-of-Experts Language Models}, 
      author={Damai Dai and Chengqi Deng and Chenggang Zhao and R. X. Xu and Huazuo Gao and Deli Chen and Jiashi Li and Wangding Zeng and Xingkai Yu and Y. Wu and Zhenda Xie and Y. K. Li and Panpan Huang and Fuli Luo and Chong Ruan and Zhifang Sui and Wenfeng Liang},
      year={2024},
      eprint={2401.06066},
      archivePrefix={arXiv},
      primaryClass={cs.CL},
      url={https://arxiv.org/abs/2401.06066}, 
}

@article{10.1162/neco.1994.6.2.181,
author = {Jordan, Michael I. and Jacobs, Robert A.},
title = {Hierarchical mixtures of experts and the EM algorithm},
year = {1994},
issue_date = {March 1994},
publisher = {MIT Press},
address = {Cambridge, MA, USA},
volume = {6},
number = {2},
issn = {0899-7667},
url = {https://doi.org/10.1162/neco.1994.6.2.181},
doi = {10.1162/neco.1994.6.2.181},
journal = {Neural Comput.},
month = mar,
pages = {181–214},
numpages = {34}
}

@article{10.1162/neco.1991.3.1.79,
    author = {Jacobs, Robert A. and Jordan, Michael I. and Nowlan, Steven J. and Hinton, Geoffrey E.},
    title = {Adaptive Mixtures of Local Experts},
    journal = {Neural Computation},
    volume = {3},
    number = {1},
    pages = {79-87},
    year = {1991},
    month = {03},
    issn = {0899-7667},
    doi = {10.1162/neco.1991.3.1.79},
    url = {https://doi.org/10.1162/neco.1991.3.1.79},
    eprint = {https://direct.mit.edu/neco/article-pdf/3/1/79/812104/neco.1991.3.1.79.pdf},
}

@misc{du2022glamefficientscalinglanguage,
      title={GLaM: Efficient Scaling of Language Models with Mixture-of-Experts}, 
      author={Nan Du and Yanping Huang and Andrew M. Dai and Simon Tong and Dmitry Lepikhin and Yuanzhong Xu and Maxim Krikun and Yanqi Zhou and Adams Wei Yu and Orhan Firat and Barret Zoph and Liam Fedus and Maarten Bosma and Zongwei Zhou and Tao Wang and Yu Emma Wang and Kellie Webster and Marie Pellat and Kevin Robinson and Kathleen Meier-Hellstern and Toju Duke and Lucas Dixon and Kun Zhang and Quoc V Le and Yonghui Wu and Zhifeng Chen and Claire Cui},
      year={2022},
      eprint={2112.06905},
      archivePrefix={arXiv},
      primaryClass={cs.CL},
      url={https://arxiv.org/abs/2112.06905}, 
}

@misc{deepseekai2024deepseekv2strongeconomicalefficient,
      title={DeepSeek-V2: A Strong, Economical, and Efficient Mixture-of-Experts Language Model}, 
      author={DeepSeek-AI},
      year={2024},
      eprint={2405.04434},
      archivePrefix={arXiv},
      primaryClass={cs.CL},
      url={https://arxiv.org/abs/2405.04434}, 
}

@misc{deepseekai2025deepseekv3technicalreport,
      title={DeepSeek-V3 Technical Report}, 
      author={DeepSeek-AI},
      year={2025},
      eprint={2412.19437},
      archivePrefix={arXiv},
      primaryClass={cs.CL},
      url={https://arxiv.org/abs/2412.19437}, 
}

@misc{wang2024auxiliarylossfreeloadbalancingstrategy,
      title={Auxiliary-Loss-Free Load Balancing Strategy for Mixture-of-Experts}, 
      author={Lean Wang and Huazuo Gao and Chenggang Zhao and Xu Sun and Damai Dai},
      year={2024},
      eprint={2408.15664},
      archivePrefix={arXiv},
      primaryClass={cs.LG},
      url={https://arxiv.org/abs/2408.15664}, 
}

@misc{zoph2022stmoedesigningstabletransferable,
      title={ST-MoE: Designing Stable and Transferable Sparse Expert Models}, 
      author={Barret Zoph and Irwan Bello and Sameer Kumar and Nan Du and Yanping Huang and Jeff Dean and Noam Shazeer and William Fedus},
      year={2022},
      eprint={2202.08906},
      archivePrefix={arXiv},
      primaryClass={cs.CL},
      url={https://arxiv.org/abs/2202.08906}, 
}

@misc{zhou2022mixtureofexpertsexpertchoicerouting,
      title={Mixture-of-Experts with Expert Choice Routing}, 
      author={Yanqi Zhou and Tao Lei and Hanxiao Liu and Nan Du and Yanping Huang and Vincent Zhao and Andrew Dai and Zhifeng Chen and Quoc Le and James Laudon},
      year={2022},
      eprint={2202.09368},
      archivePrefix={arXiv},
      primaryClass={cs.LG},
      url={https://arxiv.org/abs/2202.09368}, 
}

@misc{roller2021hashlayerslargesparse,
      title={Hash Layers For Large Sparse Models}, 
      author={Stephen Roller and Sainbayar Sukhbaatar and Arthur Szlam and Jason Weston},
      year={2021},
      eprint={2106.04426},
      archivePrefix={arXiv},
      primaryClass={cs.LG},
      url={https://arxiv.org/abs/2106.04426}, 
}

@misc{lewis2021baselayerssimplifyingtraining,
      title={BASE Layers: Simplifying Training of Large, Sparse Models}, 
      author={Mike Lewis and Shruti Bhosale and Tim Dettmers and Naman Goyal and Luke Zettlemoyer},
      year={2021},
      eprint={2103.16716},
      archivePrefix={arXiv},
      primaryClass={cs.CL},
      url={https://arxiv.org/abs/2103.16716}, 
}

\appendix

\section{Proof of Proposition~\ref{prop:ratio}}
\label{app:proof}

Write $q=1-m=\sum_{i\in\mathcal{K}}g_i$, $A=\sum_{i\in\mathcal{K}}g_i^2$ and $\mathcal{T}=\mathcal{S}\setminus\mathcal{K}$, so that $R=A/\big(q^2\sum_{i\in\mathcal{S}}g_i^2\big)$ by Eq.~(\ref{eq:ratio}). We assume positive gates, which holds for softmax and sigmoid routers.

We first derive the general bounds. The renormalized gates $\tilde{g}$ sum to one over $\mathcal{K}$,
\[
  \sum_{i\in\mathcal{K}}\tilde{g}_i = 1 .
\]
By the Cauchy--Schwarz inequality,
\[
  1 = \Big(\sum_{i\in\mathcal{K}} 1\cdot\tilde{g}_i\Big)^2
    \le \Big(\sum_{i\in\mathcal{K}} 1\Big)\Big(\sum_{i\in\mathcal{K}}\tilde{g}_i^2\Big)
    = |\mathcal{K}|\sum_{i\in\mathcal{K}}\tilde{g}_i^2 ,
\]
\[
  \frac{1}{|\mathcal{K}|} \le \sum_{i\in\mathcal{K}}\tilde{g}_i^2 \le 1 .
\]
Likewise,
\[
  \frac{1}{k_0} \le \sum_{i\in\mathcal{S}}g_i^2 \le 1 .
\]
Since $|\mathcal{K}|\le k_0$, the ratio satisfies
\[
  \frac{1}{k_0} \le R \le k_0 .
\]

Next, consider order-consistent masks. Let
\[
  \gamma = \min_{i\in\mathcal{K}} g_i .
\]
Then
\[
  A = \sum_{i\in\mathcal{K}} g_i^2 \ge \gamma \sum_{i\in\mathcal{K}} g_i = \gamma q .
\]
Order consistency gives $g_j\le\gamma$ for every $j\in\mathcal{T}$, so
\begin{equation}
  \sum_{j\in\mathcal{T}}g_j^2 \le \gamma\sum_{j\in\mathcal{T}}g_j = \gamma m \le \frac{A\,m}{q}.
  \label{eq:proof-tail}
\end{equation}
Hence
\[
  \sum_{i\in\mathcal{S}}g_i^2 = A+\sum_{j\in\mathcal{T}}g_j^2 \le \frac{A(q+m)}{q} = \frac{A}{q}.
\]
Substituting this bound into $R$,
\begin{equation}
  R = \frac{A}{q^2\sum_{i\in\mathcal{S}}g_i^2} \ge \frac{A}{q\,A} = \frac{1}{1-m} \ge 1.
  \label{eq:proof-lower}
\end{equation}
If $R=1$, then $q=1$, i.e., $m=0$; with positive gates this means $\mathcal{T}=\emptyset$.

\section{Derivation of $U_b$ and $Z_b$}
\label{app:sums}

We derive Eq.~(\ref{eq:bucket-from-sums}) for one layer and one skipping rule. Let $j=1,\dots,d$ index the dimensions, so that $y_{f,j}$, $\mu_{f,j}$, $\sigma_{f,j}$ and $S^{ff}_{b,j}$ are the $j$-th entries of $\mathbf{y}_f$, $\boldsymbol{\mu}_f$, $\boldsymbol{\sigma}_f$ and $\mathbf{S}^{ff}_b$. Unless stated otherwise, sums run over the tokens $\mathbf{x}$ of the bucket $\mathcal{B}_b$.

\paragraph{Derivation of $U_b$.}
By Eq.~(\ref{eq:std}), $u_j=(y_{f,j}-\mu_{f,j})/\sigma_{f,j}$. The moments $\mu_{f,j}$ and $\sigma_{f,j}$ are computed on the whole calibration set, so they are constants inside the bucket and can be taken out of the sum:
\begin{equation}
  \begin{aligned}
    \sum u_j^2
    &= \frac{1}{\sigma_{f,j}^2}\sum\big(y_{f,j}-\mu_{f,j}\big)^2 \\
    &= \frac{1}{\sigma_{f,j}^2}\Big(\textstyle\sum y_{f,j}^2-2\mu_{f,j}\sum y_{f,j} \\
    &\qquad\qquad+n_b\mu_{f,j}^2\Big) \\
    &= \frac{S^{ff}_{b,j}-2\mu_{f,j}S^{f}_{b,j}+n_b\mu_{f,j}^2}{\sigma_{f,j}^2}.
  \end{aligned}
  \label{eq:app-u}
\end{equation}
Averaging over the $n_b$ tokens of the bucket and over the $d$ dimensions gives
\begin{equation}
  \begin{aligned}
    U_b &= \frac{1}{d}\sum_{j=1}^{d}\frac{1}{n_b}\sum u_j^2 \\
        &= \frac{1}{d}\sum_{j=1}^{d}\frac{S^{ff}_{b,j}-2\mu_{f,j}S^{f}_{b,j}+n_b\mu_{f,j}^2}{n_b\,\sigma_{f,j}^2},
  \end{aligned}
  \label{eq:app-ub}
\end{equation}
which is the first line of Eq.~(\ref{eq:bucket-from-sums}) with $\langle\cdot\rangle_d=\frac{1}{d}\sum_{j}$.

\paragraph{Derivation of $Z_b$.}
By Eq.~(\ref{eq:lda}), $a_j=\sigma_{f,j}/\sigma_{p,j}$, so $z_j=a_j(y_{p,j}-\mu_{p,j})/\sigma_{f,j}=(y_{p,j}-\mu_{p,j})/\sigma_{p,j}$. The same expansion with $(y_{p,j},\mu_{p,j},\sigma_{p,j})$ in place of $(y_{f,j},\mu_{f,j},\sigma_{f,j})$ gives
\begin{equation}
  Z_b = \frac{1}{d}\sum_{j=1}^{d}\frac{S^{pp}_{b,j}-2\mu_{p,j}S^{p}_{b,j}+n_b\mu_{p,j}^2}{n_b\,\sigma_{p,j}^2},
  \label{eq:app-zb}
\end{equation}
the second line of Eq.~(\ref{eq:bucket-from-sums}). Both statistics, and hence $c_b^\star$ in Eq.~(\ref{eq:cstar}), require only $n_b$, $\mathbf{S}^{f}_b$, $\mathbf{S}^{p}_b$, $\mathbf{S}^{ff}_b$, $\mathbf{S}^{pp}_b$ and the global moments.

\section{Skipping Rules}
\label{app:rules}

The following formally describes the five skipping rules. Every rule operates on the original top-$k_0$ set $\mathcal{S}(\mathbf{x})$ of Section~\ref{sec:prelim} and returns a keep mask $\kappa$; dropped slots are not refilled, and the kept gates are renormalized by Eq.~(\ref{eq:renorm}).

\paragraph{Notation.}
We omit the layer index $\ell$ when a quantity refers to a single MoE layer, and let $L$ be the number of MoE layers. For a token $\mathbf{x}$, let $g_{(1)}\ge\dots\ge g_{(k_0)}$ be its gates of Eq.~(\ref{eq:gates}) sorted in decreasing order, where $(r)\in\mathcal{S}$ is the slot of rank $r$, and let
\begin{equation}
  P_r = \sum_{r'<r} g_{(r')}, \qquad r=1,\dots,k_0,
  \label{eq:app-prefix}
\end{equation}
be the gate mass ranked above slot $(r)$, so that $P_1=0$. $\mathcal{X}_{\mathrm{cal}}$ is a set of calibration tokens, and the batch-level rules act on a batch $\mathcal{M}$ of $M$ tokens processed by the layer at once.

\paragraph{ACE.}
ACE~\citep{xu2026aceadaptivecalibrationfreeexpert} combines the gates with two weight-only amplitude scores per expert. For a row vector $\mathbf{x}$, a SwiGLU expert computes $E_e(\mathbf{x})=\big(\operatorname{SiLU}(\mathbf{x}\mathbf{W}^{\mathrm{g}}_e)\odot\mathbf{x}\mathbf{W}^{\mathrm{u}}_e\big)\mathbf{W}^{\mathrm{d}}_e$ with $\mathbf{W}^{\mathrm{g}}_e,\mathbf{W}^{\mathrm{u}}_e\in\mathbb{R}^{d\times d_{\mathrm{ff}}}$ and $\mathbf{W}^{\mathrm{d}}_e\in\mathbb{R}^{d_{\mathrm{ff}}\times d}$. Let $\boldsymbol{\Gamma}$ be the diagonal matrix of the RMSNorm weights before the layer and $\mathbf{w}_e$ the router weight vector of expert $e$. The global spectral proxy (GSP) and the router-conditioned refinement (RCR) are
\begin{equation}
  \begin{aligned}
    \tilde{a}^{\mathrm{G}}_e &= \Big(\|\boldsymbol{\Gamma}\mathbf{W}^{\mathrm{u}}_e\|_F\,\|\boldsymbol{\Gamma}\mathbf{W}^{\mathrm{g}}_e\mathbf{W}^{\mathrm{d}}_e\|_F \\
    &\qquad\cdot\|\boldsymbol{\Gamma}\mathbf{W}^{\mathrm{g}}_e\|_F\,\|\boldsymbol{\Gamma}\mathbf{W}^{\mathrm{u}}_e\mathbf{W}^{\mathrm{d}}_e\|_F\Big)^{1/2}, \\
    \tilde{a}^{\mathrm{R}}_e &= \frac{\|E_e(\mathbf{q}_e)\|_2}{\|\mathbf{q}_e\|_2}, \qquad \mathbf{q}_e=\frac{\mathbf{w}_e}{\operatorname{RMS}(\mathbf{w}_e)},
  \end{aligned}
  \label{eq:app-ace-amp}
\end{equation}
and each is normalized by its layer mean, $A^{\mathrm{G}}_e=\tilde{a}^{\mathrm{G}}_e/\big(\frac{1}{N}\sum_{j=1}^{N}\tilde{a}^{\mathrm{G}}_j\big)$, and likewise $A^{\mathrm{R}}_e$. Both tables are computed once from the weights. At inference, the slot scores are
\begin{equation}
  \begin{aligned}
    p^{\mathrm{G}}_i &= \frac{g_iA^{\mathrm{G}}_i}{\sum_{j\in\mathcal{S}}g_jA^{\mathrm{G}}_j}, \qquad
    p^{\mathrm{R}}_i = \frac{g_iA^{\mathrm{R}}_i}{\sum_{j\in\mathcal{S}}g_jA^{\mathrm{R}}_j}, \\
    c_i &= \max\big(p^{\mathrm{G}}_i,\ \omega\,p^{\mathrm{R}}_i\big), \qquad i\in\mathcal{S}.
  \end{aligned}
  \label{eq:app-ace-score}
\end{equation}
With $c_{[1]}\ge\dots\ge c_{[k_0]}$ the sorted scores, ACE keeps the shortest prefix whose score mass reaches $1-\tau$ of the total,
\begin{equation}
  \kappa_{[r]} = \mathbf{1}\Big[\textstyle\sum_{r'<r}c_{[r']} < (1-\tau)\sum_{j\in\mathcal{S}}c_j\Big],
  \label{eq:app-ace-keep}
\end{equation}
and always keeps the slot with the largest gate $g_{(1)}$.
\emph{Setting.} We use $\tau=0.7$ and $\omega=0.1$. ACE needs no calibration data.

\paragraph{Top-$\boldsymbol{p}$.}
Following the top-$p$ routing of \citet{huang2024hardertasksneedexperts}, the rule keeps the shortest prefix of the sorted gates whose mass reaches $p$,
\begin{equation}
  \kappa_{(r)} = \mathbf{1}\big[P_r < p\big],
  \label{eq:app-topp}
\end{equation}
so the top-1 slot is always kept. \citet{huang2024hardertasksneedexperts} apply the criterion over all $N$ experts during training; we apply it training-free within $\mathcal{S}$, with the gates normalized over the top-$k_0$ set.
\emph{Setting.} We use $p=0.3$. The rule needs no calibration data.

\paragraph{Alloc-T.}
Alloc-T~\citep{liu2026allocmoebudgetawareexpertactivation} works in two stages. It first assigns each layer a budget $K_\ell\in\{1,\dots,k_0\}$ by sensitivity profiling. Layers are profiled from the deepest to the shallowest: when layer $\ell$ is profiled, all shallower layers use $k_0$ and all deeper layers use $1$, and we record the calibration perplexity $\mathrm{PPL}_\ell(k)$ for $K_\ell=k_0,\dots,1$. With the normalized sensitivity $\hat{S}_\ell(k)=\mathrm{PPL}_\ell(k)/\mathrm{PPL}_\ell(k_0)$ and a target average budget $\bar{k}$ per token and layer, the budgets solve
\begin{equation}
  \begin{aligned}
    \min_{K_1,\dots,K_L}\ &\sum_{\ell=1}^{L}\hat{S}_\ell(K_\ell) \\
    \text{s.t.}\ &\sum_{\ell=1}^{L}K_\ell\le\bar{k}L,\quad 1\le K_\ell\le k_0,
  \end{aligned}
  \label{eq:app-alloc-dp}
\end{equation}
which a dynamic program over layers solves exactly with the recurrence $D[\ell,b]=\min_{1\le k\le\min(k_0,b)}D[\ell-1,b-k]+\hat{S}_\ell(k)$ and $D[0,0]=0$.
It then uses $K_\ell$ as the average budget of a batch and redistributes the slots of the batch across its tokens. With binary variables $z_{\mathbf{x},i}$ over the slots $i\in\mathcal{S}(\mathbf{x})$ of the tokens $\mathbf{x}\in\mathcal{M}$, it solves
\begin{equation}
  \begin{aligned}
    \max_{z}\ &\sum_{\mathbf{x}\in\mathcal{M}}\sum_{i\in\mathcal{S}(\mathbf{x})}z_{\mathbf{x},i}\,g_i(\mathbf{x}) \\
    \text{s.t.}\ &\sum_{\mathbf{x},i}z_{\mathbf{x},i}\le K_\ell M,\quad \sum_{i}z_{\mathbf{x},i}\ge K_{\mathrm{base}}\ \ \forall\mathbf{x},
  \end{aligned}
  \label{eq:app-alloct}
\end{equation}
and sets $\kappa_i(\mathbf{x})=z_{\mathbf{x},i}$. The solution keeps the top-$K_{\mathrm{base}}$ slots of every token and then the $(K_\ell-K_{\mathrm{base}})M$ largest gates among the remaining $(k_0-K_{\mathrm{base}})M$ slots of the batch.
\emph{Setting.} $\bar{k}=2$ and $K_{\mathrm{base}}=1$. The perplexities of the first stage are measured on 4 sequences of 2,048 tokens from the C4 training split. During calibration we use batches of $M=32$ tokens.

\paragraph{XShare.}
XShare~\citep{vankov2026xsharecollaborativeinbatchexpert} restricts all tokens of a batch to a shared expert set $\mathcal{E}$. The batch score of expert $e$ and the warm-up set are
\begin{equation}
  G_e = \sum_{\mathbf{x}\in\mathcal{M}} g_e(\mathbf{x}), \qquad \mathcal{W} = \bigcup_{\mathbf{x}\in\mathcal{M}}\{e_{(1)}(\mathbf{x})\},
  \label{eq:app-xshare-score}
\end{equation}
where $g_e(\mathbf{x})=0$ for $e\notin\mathcal{S}(\mathbf{x})$ and $e_{(1)}(\mathbf{x})$ is the expert in the top-1 slot of $\mathbf{x}$. Starting from $\mathcal{E}=\mathcal{W}$, the greedy step adds $\arg\max_{e\notin\mathcal{E}}G_e$ to $\mathcal{E}$ until
\begin{equation}
  \textstyle\sum_{e\in\mathcal{E}}G_e \ge \eta\sum_{e=1}^{N}G_e = \eta M,
  \label{eq:app-xshare-stop}
\end{equation}
and the keep mask is $\kappa_i(\mathbf{x})=\mathbf{1}[i\in\mathcal{E}]$. Because $\mathcal{E}\supseteq\mathcal{W}$, every token keeps its top-1 slot. We do not apply the refinement step of XShare, which re-routes each token to its top-$k$ experts within $\mathcal{E}$, since it would refill dropped slots.
\emph{Setting.} One threshold $\eta$ is shared by all layers and found by bisection so that the average kept count is $\bar{k}=2$ when 8 sequences of 2,048 tokens from the C4 training split are randomly shuffled and partitioned into batches of $M=32$ tokens.

\paragraph{REAP.}
REAP~\citep{lasby2026reap} is a static rule that removes the same experts for every token. The saliency of expert $e$ is its router-weighted activation norm averaged over the calibration tokens routed to it,
\begin{equation}
  S_e = \frac{1}{|\mathcal{X}_e|}\sum_{\mathbf{x}\in\mathcal{X}_e}g_e(\mathbf{x})\,\|E_e(\mathbf{x})\|_2,
  \label{eq:app-reap}
\end{equation}
with $\mathcal{X}_e=\{\mathbf{x}\in\mathcal{X}_{\mathrm{cal}}:e\in\mathcal{S}(\mathbf{x})\}$. In each layer, the set $\mathcal{P}_\ell$ of the $\lfloor\rho N\rfloor$ experts with the lowest $S_e$ is removed. Unlike REAP, which re-selects the top-$k_0$ experts among the remaining ones, we keep the original routing and drop the slots of removed experts, $\kappa_i=\mathbf{1}[i\notin\mathcal{P}_\ell]$; if all slots of a token point to removed experts, the top-1 slot is kept.
\emph{Setting.} $\rho=0.5$; $\mathcal{X}_{\mathrm{cal}}$ is 8 sequences of 2,048 tokens from evol-codealpaca formatted with the chat template of the model.

\section{Evaluation Details}
\label{app:eval}

We report pass@1 for code and accuracy for all other benchmarks, where the IFEval score is the mean of the prompt-level and instruction-level accuracies under strict and loose evaluation. All benchmarks are run zero-shot with EvalScope \citep{evalscope_2024} through the OpenAI-compatible API of vLLM. Decoding uses temperature 1.0 and top-$p$ 0.95. Each benchmark is run 8 times with seeds 0--7, and we report the mean over the runs. For MMLU-Pro we sample 100 questions from each of its 14 subsets, and for SuperGPQA 30 questions from each of its 72 subsets. LiveCodeBench uses the v5\_v6 subset, and GPQA-Diamond, MATH-500, AIME25, AIME26, HumanEval+ and IFEval are used in full. The maximum number of generated tokens is 4,096 for HumanEval+, 8,192 for IFEval, and 16,384 for the other benchmarks. Thinking is enabled only on AIME25 and AIME26 for Qwen3.6-35B-A3B, Qwen3.5-122B-A10B-FP8 and GLM-4.7-Flash, where the budget is raised to 81,920 tokens; the other models have no thinking mode, and thinking is disabled on all other benchmarks. Code is executed in a Docker sandbox without network access.

\begin{figure*}[!t]
    \centering
    \includegraphics[width=1.0\linewidth]{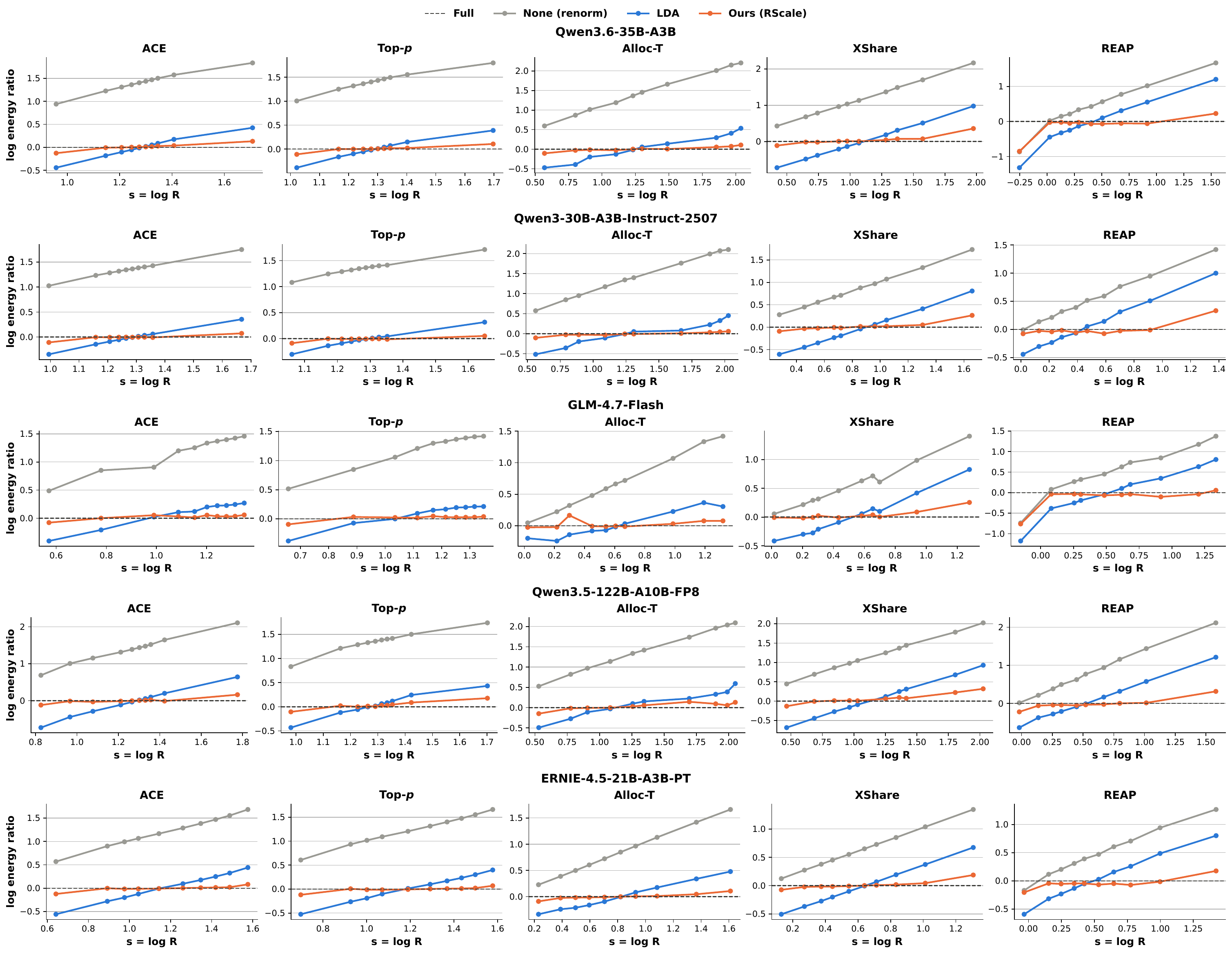}
    \caption{Output energy ratio versus the routing signal $s=\log R$ for all five models (rows) and five skipping rules (columns), computed as in Figure~\ref{case4b}.}
    \label{energy_all}
\end{figure*}

\begin{figure*}[!t]
    \centering
    \includegraphics[width=0.85\linewidth]{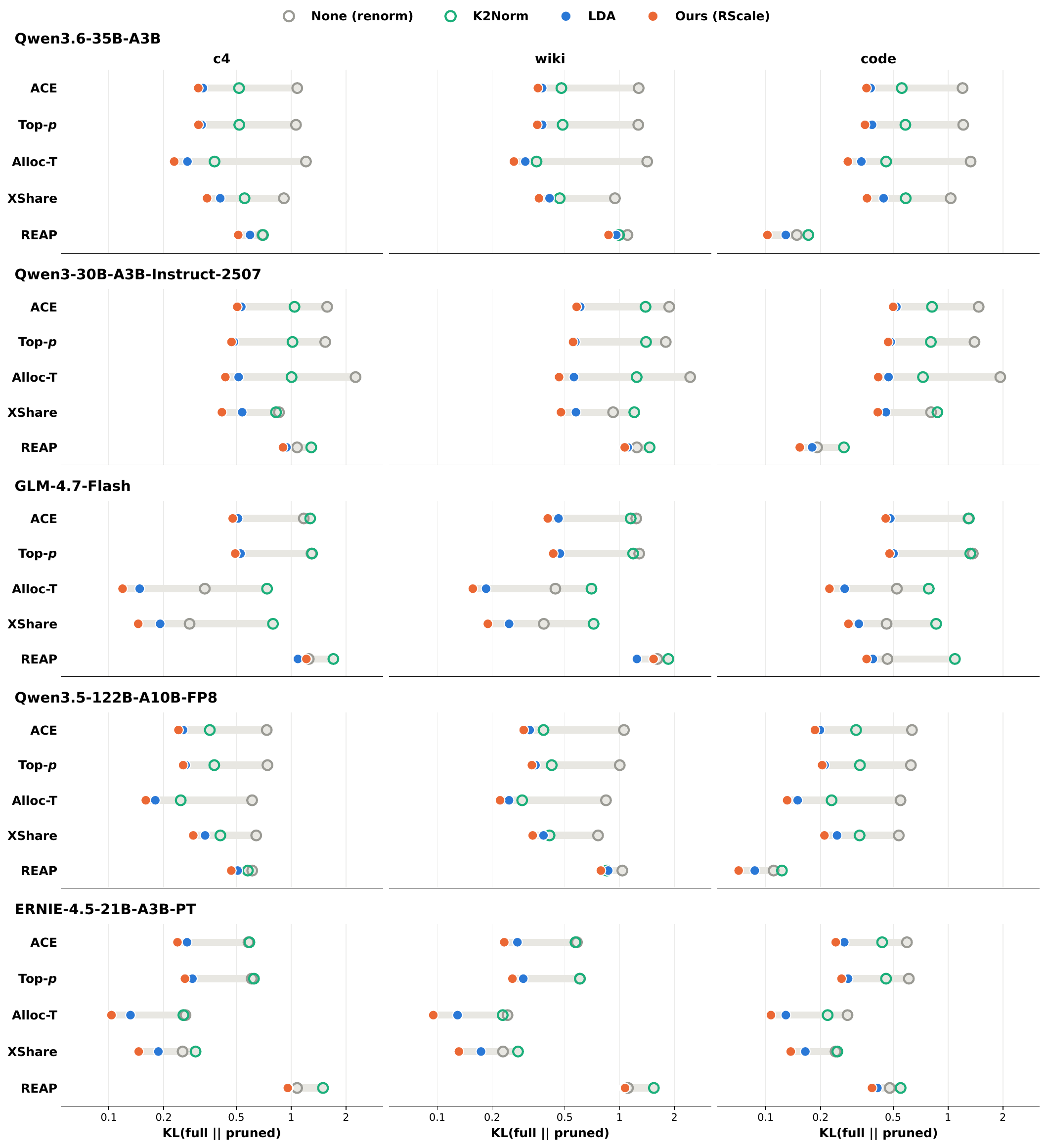}
    \caption{KL divergence from the full to the pruned next-token distribution (log scale, lower is better) for all five models, computed as in Figure~\ref{fig:kl}. All panels share the same axis.}
    \label{fig:kl-all}
\end{figure*}

\begin{figure*}[!t]
    \centering
    \includegraphics[width=1.0\linewidth]{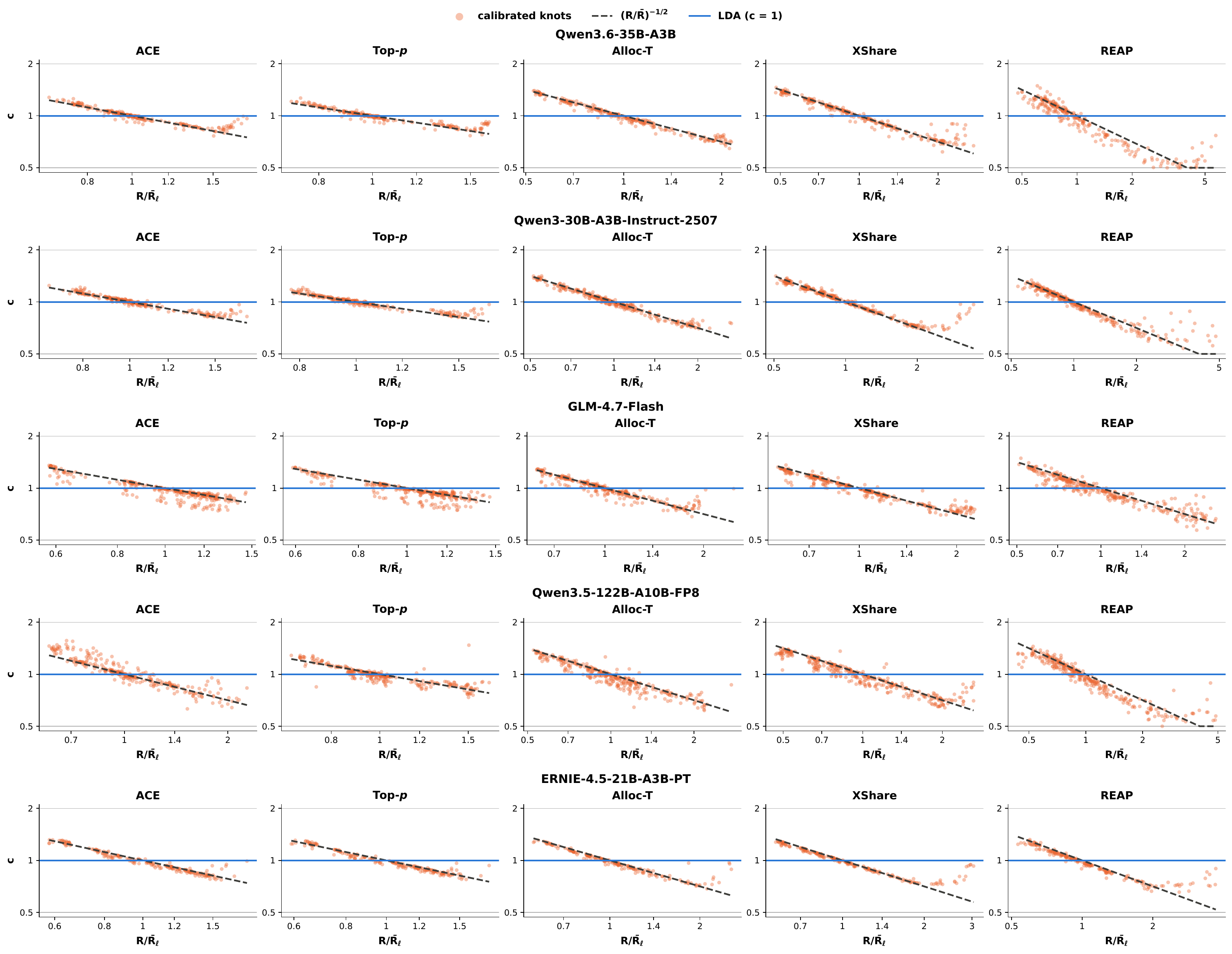}
    \caption{Calibrated knots of \ours{} against $R/\bar{R}_\ell$ for all five models (rows) and five skipping rules (columns), plotted as in Figure~\ref{fig:c}.}
    \label{fig:c-all}
\end{figure*}

\section{Ablation}
\label{sec:ablation}

All ablations use Qwen3.6-35B-A3B and four benchmarks, MATH-500, GPQA-Diamond, HumanEval+ and IFEval, with the protocol of Section~\ref{sec:setup}. Each cell is averaged over the skipping rules and over 8 seeds. Appendix~\ref{app:ablation} breaks the results down by rule.

\begin{table}[t]
  \centering
  \resizebox{\columnwidth}{!}{%
  \begin{tabular}{lccccc}
    \toprule
    \textbf{Correction} & MATH-500 & GPQA-D & HumanEval+ & IFEval & \textbf{Avg.} \\
    \midrule
    Full & 98.10 & 83.58 & 91.31 & 86.37 & 89.84 \\
    \midrule
    LDA ($c_\ell\equiv 1$) & 89.31 & 60.48 & 70.97 & 74.96 & 73.93 \\
    Ours ($B$=8, $\lambda$=100) & \textbf{91.99} & 64.91 & 76.98 & 76.82 & 77.68 \\
    \midrule
    Closed form & 90.50 & 63.53 & 74.75 & 75.14 & 75.98 \\
    Closed form w/o calib & 90.80 & 61.21 & 72.93 & 74.12 & 74.76 \\
    \midrule
    $\lambda$=10 & 91.94 & \textbf{65.32} & \textbf{77.85} & \textbf{77.24} & \textbf{78.09} \\
    $\lambda$=1000 & 90.98 & 63.74 & 76.67 & 76.34 & 76.93 \\
    $\lambda$=$10^4$ & 90.00 & 61.06 & 73.05 & 74.97 & 74.77 \\
    \midrule
    $B$=2 & 90.75 & 63.65 & 75.93 & 76.90 & 76.81 \\
    $B$=4 & 91.25 & 63.81 & 75.78 & 76.83 & 76.92 \\
    $B$=16 & 91.50 & 64.87 & 77.50 & 76.74 & 77.65 \\
    \bottomrule
  \end{tabular}}
  \caption{Ablation of the scale function on Qwen3.6-35B-A3B, averaged over the five skipping rules and 8 seeds. Closed form uses $c=(R/\bar{R}_\ell)^{-1/2}$ without buckets and shrinkage, and Closed form w/o calib applies $\hat{\mathbf{y}}=R^{-1/2}\mathbf{y}_p$ directly without LDA. The $\lambda$ and $B$ rows change one hyperparameter of \ours{} at a time. Bold marks the best correction in each column. Scores in \%.}
  \label{tab:abl-rscale}
\end{table}

\paragraph{Closed form.}
The first variant in Table~\ref{tab:abl-rscale} replaces the calibrated knots by the prediction of Section~\ref{sec:lda}. For every token it computes $R$ from the gates and applies $c=(R/\bar{R}_\ell)^{-1/2}$, where $\bar{R}_\ell$ is measured on the calibration tokens of each layer and rule; it uses no buckets and no shrinkage. It scores 75.98 on average, 2.05 points above LDA, and it is 1.70 points below \ours{}. Without any fitting, the closed form already recovers more than half of the gain of \ours{} (2.05 of 3.75 points), so most of the gain comes from the token-dependent rescaling predicted in Section~\ref{sec:lda}. The calibrated knots follow this prediction, and fitting them adds another 1.7 points. The second variant also drops the calibration statistics and restores the energy of each token directly from Eq.~(\ref{eq:ratio}): it multiplies the renormalized gates by $R^{-1/2}$, i.e., $\hat{\mathbf{y}}=R^{-1/2}\mathbf{y}_p$, so that $\|\hat{\mathbf{y}}\|^2\approx\|\mathbf{y}_f\|^2$ without any calibration data. It differs from the closed form in that the calibrated $\boldsymbol{\mu}_f$, $\boldsymbol{\mu}_p$ and $\mathbf{a}$ of LDA are not used. It scores 74.76 on average, only 0.83 points above LDA and 1.22 points below the closed form, and it falls below LDA under Top-$p$, Alloc-T and REAP (Table~\ref{tab:abl-rscale-rule}). The token-dependent factor thus works best on top of the calibrated mean and variance of LDA. \ours{} therefore uses the knots fitted by bucket-wise moment matching (Section~\ref{sec:fit}) instead of the closed form: they are set from the energy ratio measured in each bucket rather than from the idealized energy model, so they also correct the buckets where this ratio departs from the prediction.

\paragraph{Shrinkage.}
With 8,192 calibration tokens per layer and $B=8$, a bucket holds about $n/B= 1{,}024$ tokens. The weight $n_b/(n_b+\lambda)$ of Eq.~(\ref{eq:shrink}) is then about 0.99, 0.91, 0.51 and 0.09 for $\lambda=10$, 100, 1000 and $10^4$. The gain over LDA falls with this weight, from 4.16 points at $\lambda=10$ and 3.75 at $\lambda=100$ to 3.00 at $\lambda=1000$ and 0.84 at $\lambda=10^4$, and the last setting is close to LDA, as expected from $\lambda\to\infty$. Weak shrinkage does not hurt, so the buckets are large enough for $c_b^\star$ to be estimated reliably.

\paragraph{Number of buckets.}
With $B=2$, $c_\ell$ is a single linear segment in $s$ between two knots, and it already recovers 2.88 of the 3.75 points that \ours{} gains over LDA with $B=8$. The average rises to 76.92 with $B=4$ and saturates at $B=8$ (77.68), with no further gain at $B=16$ (77.65). Every $B$ we tried improves over LDA on all four benchmarks, and per rule (Table~\ref{tab:abl-rscale-rule}) it raises the average over LDA under all five skipping rules, by 1.16 to 5.86 points. \ours{} therefore does not depend on a finely tuned number of buckets, and even the coarsest setting $B=2$ keeps most of its gain.

\begin{table}[t]
  \centering
  \resizebox{\columnwidth}{!}{%
  \begin{tabular}{llccccc}
    \toprule
    \textbf{Calib.} & \textbf{Corr.} & MATH-500 & GPQA-D & HumanEval+ & IFEval & \textbf{Avg.} \\
    \midrule
    \multirow{2}{*}{C4, 2K} & LDA & 87.44 & 67.11 & 66.35 & 72.65 & 73.39 \\
      & Ours & 89.50 & 70.93 & 73.29 & 75.08 & \textbf{77.20} \\
    \midrule
    \multirow{2}{*}{C4, 8K} & LDA & 87.63 & 66.35 & 66.25 & 73.61 & 73.46 \\
      & Ours & 90.69 & 70.80 & 73.15 & 75.59 & \textbf{77.56} \\
    \midrule
    \multirow{2}{*}{C4, 32K} & LDA & 87.75 & 66.98 & 68.62 & 73.89 & 74.31 \\
      & Ours & 90.52 & 71.02 & 73.86 & 75.51 & \textbf{77.73} \\
    \midrule
    \multirow{2}{*}{Math, 8K} & LDA & 84.40 & 64.21 & 65.85 & 73.58 & 72.01 \\
      & Ours & 87.36 & 68.67 & 72.37 & 75.37 & \textbf{75.94} \\
    \midrule
    \multirow{2}{*}{Code, 8K} & LDA & 86.47 & 64.84 & 64.65 & 72.13 & 72.02 \\
      & Ours & 88.91 & 69.56 & 72.52 & 75.35 & \textbf{76.59} \\
    \bottomrule
  \end{tabular}}
  \caption{Effect of the calibration data on Qwen3.6-35B-A3B, averaged over four skipping rules (all but REAP) and 8 seeds. C4, 8K is the setting of Table~\ref{tab:main}. Bold marks the better correction on Avg.\ for each calibration set. Scores in \%.}
  \label{tab:abl-calib}
\end{table}

\paragraph{Calibration size.}
Table~\ref{tab:abl-calib} recalibrates LDA and \ours{} on 1 and 16 sequences of 2,048 tokens from C4, i.e., 2K and 32K tokens, while the skipping rules keep the parameters of the main experiments. We exclude REAP here for the reason given in Section~\ref{sec:setup}. Both corrections are insensitive to the amount of data. From 2K to 32K tokens, \ours{} improves by 0.53 points and LDA by 0.92 points, and \ours{} calibrated on 2K tokens is still 2.89 points above LDA calibrated on 32K tokens. The gain over LDA is between 3.42 and 4.10 points for the three sizes. 

\paragraph{Calibration domain.}
We also calibrate on 8K tokens of math web text from OpenWebMath \citep{paster2023openwebmath} and of code from evol-codealpaca \citep{luo2023wizardcoder}. With math and code calibration, \ours{} averages 75.94 and 76.59 and LDA 72.01 and 72.02, compared with 77.56 and 73.46 on C4. Both corrections lose between 0.97 and 1.62 points relative to C4, while the gain of \ours{} over LDA stays between 3.93 and 4.57 points on C4, math and code, so the advantage of \ours{} does not depend on the domain of the calibration data.

\section{Statistical Significance}
\label{app:significance}

We test whether the gain of \ours{} over LDA in Table~\ref{tab:main} can be explained by sampling noise. We use a paired $t$-test over the evaluation seeds. Both corrections are evaluated on the same questions with the same seeds 0--7, so the runs are paired by seed.

For one model and benchmark, let $X^{\mathrm{ours}}_s$ and $X^{\mathrm{lda}}_s$ be the scores with seed $s$, each averaged over the five skipping rules as in Table~\ref{tab:main}; for Avg., the score is also averaged over the nine benchmarks. With the $n=8$ paired differences $d_s=X^{\mathrm{ours}}_s-X^{\mathrm{lda}}_s$, their mean $\bar{d}$ and standard deviation $s_d$, the null hypothesis $H_0:\mathbb{E}[d]=0$ is tested with
\begin{equation}
  t = \frac{\bar{d}}{s_d/\sqrt{n}},
  \label{eq:app-ttest}
\end{equation}
which follows a Student $t$-distribution with $n-1=7$ degrees of freedom under $H_0$. We report the two-sided $p$-value and the 95\% confidence interval $\bar{d}\pm t_{0.975,7}\,s_d/\sqrt{n}$ of the expected gain, with $t_{0.975,7}=2.365$. Pairing removes the variation that a seed causes in both corrections alike, which an unpaired comparison of the two means would count as noise.

Table~\ref{tab:significance} lists the results. The gain in average score is significant on all five models ($p\le 0.002$), and \ours{} scores higher than LDA with every one of the 8 seeds on every model. Per benchmark, the gain is significant in 36 of the 45 model and benchmark pairs, and in 30 after a Holm--Bonferroni correction over the 45 pairs. Five of the nine remaining pairs are AIME25 and AIME26, which have only 30 problems each, so that a single problem changes the score by 3.33 points and 8 seeds are too few to separate gains of 1 to 2.5 points from noise. A Wilcoxon signed-rank test, which does not assume normally distributed differences, gives the same decision at $p<0.05$ in all 50 cells.

\begin{table*}[t]
  \centering
  \resizebox{\textwidth}{!}{%
  \begin{tabular}{llcccccccccc}
    \toprule
    \multirow{2}{*}{\textbf{Model}} & \multirow{2}{*}{\textbf{Stat.}} & \multicolumn{3}{c}{\textbf{Math}} & \multicolumn{3}{c}{\textbf{Knowledge \& Reasoning}} & \multicolumn{2}{c}{\textbf{Code}} & \textbf{Inst.} & \multirow{2}{*}{\textbf{Avg.}} \\
    \cmidrule(lr){3-5}\cmidrule(lr){6-8}\cmidrule(lr){9-10}\cmidrule(lr){11-11}
    & & MATH-500 & AIME25 & AIME26 & MMLU-Pro & GPQA-D & SuperGPQA & HumanEval+ & LCB & IFEval & \\
    \midrule
    \multirow{3}{*}{Qwen3.6-35B-A3B}
      & $\Delta$ & 2.68 & 4.50 & 5.34 & 2.03 & 4.43 & 2.73 & 6.01 & 4.00 & 1.86 & 3.73 \\
      & 95\% CI & [1.96, 3.41] & [3.02, 5.98] & [2.67, 8.00] & [1.78, 2.28] & [3.92, 4.94] & [2.37, 3.09] & [4.43, 7.58] & [3.14, 4.87] & [1.23, 2.50] & [3.28, 4.18] \\
      & $p$ & $<$0.001 & $<$0.001 & 0.002 & $<$0.001 & $<$0.001 & $<$0.001 & $<$0.001 & $<$0.001 & $<$0.001 & $<$0.001 \\
    \midrule
    \multirow{3}{*}{Qwen3.5-122B-A10B-FP8}
      & $\Delta$ & 1.11 & 2.42 & 1.67 & 1.02 & 2.56 & 1.21 & 2.02 & 6.74 & 0.47 & 2.14 \\
      & 95\% CI & [0.49, 1.74] & [$-$1.41, 6.24] & [$-$3.58, 6.91] & [0.62, 1.41] & [1.61, 3.52] & [0.86, 1.58] & [0.99, 3.04] & [5.57, 7.91] & [$-$0.02, 0.95] & [1.07, 3.20] \\
      & $p$ & 0.004 & 0.179 & 0.477 & $<$0.001 & $<$0.001 & $<$0.001 & 0.002 & $<$0.001 & 0.059 & 0.002 \\
    \midrule
    \multirow{3}{*}{Qwen3-30B-A3B-2507}
      & $\Delta$ & 1.46 & 3.50 & 4.74 & 1.86 & 2.16 & 1.98 & 4.93 & 3.20 & 1.50 & 2.81 \\
      & 95\% CI & [0.89, 2.03] & [0.79, 6.21] & [2.04, 7.45] & [1.56, 2.15] & [1.14, 3.18] & [1.52, 2.45] & [4.12, 5.72] & [2.50, 3.89] & [0.92, 2.09] & [2.28, 3.35] \\
      & $p$ & $<$0.001 & 0.019 & 0.004 & $<$0.001 & 0.002 & $<$0.001 & $<$0.001 & $<$0.001 & $<$0.001 & $<$0.001 \\
    \midrule
    \multirow{3}{*}{GLM-4.7-Flash}
      & $\Delta$ & 2.77 & 2.17 & 2.25 & 2.72 & 0.80 & 2.01 & 1.73 & $-$0.17 & 1.04 & 1.70 \\
      & 95\% CI & [2.09, 3.44] & [0.46, 3.87] & [$-$1.79, 6.29] & [2.26, 3.17] & [$-$0.31, 1.90] & [1.67, 2.35] & [0.14, 3.34] & [$-$1.29, 0.94] & [0.26, 1.83] & [1.30, 2.10] \\
      & $p$ & $<$0.001 & 0.020 & 0.229 & $<$0.001 & 0.132 & $<$0.001 & 0.037 & 0.719 & 0.016 & $<$0.001 \\
    \midrule
    \multirow{3}{*}{ERNIE-4.5-21B-A3B-PT}
      & $\Delta$ & 3.12 & 1.17 & 1.25 & 1.80 & 5.09 & 1.62 & 3.64 & 0.60 & 1.81 & 2.23 \\
      & 95\% CI & [1.99, 4.26] & [$-$0.93, 3.27] & [$-$1.03, 3.53] & [1.36, 2.24] & [4.01, 6.17] & [1.20, 2.04] & [2.51, 4.77] & [$-$0.46, 1.66] & [1.51, 2.13] & [1.88, 2.59] \\
      & $p$ & $<$0.001 & 0.231 & 0.235 & $<$0.001 & $<$0.001 & $<$0.001 & $<$0.001 & 0.224 & $<$0.001 & $<$0.001 \\
    \bottomrule
  \end{tabular}}
    \caption{Paired $t$-test of \ours{} against LDA for each cell of Table~\ref{tab:main}. For every seed, the scores of both corrections are first averaged over the five skipping rules; $\Delta$ is the mean gain $\bar{d}$ of \ours{} over LDA across the 8 seeds, 95\% CI its confidence interval, and $p$ the two-sided $p$-value with 7 degrees of freedom. $\Delta$ and CI are in points (\%).}
    \label{tab:significance}
\end{table*}

\section{Results per Skipping Rule}
\label{app:per-rule}

Tables~\ref{tab:rule-ace}--\ref{tab:rule-reap} break down Table~\ref{tab:main} by skipping rule. Each cell is the mean over 8 seeds, and Avg.\ is the mean over the nine benchmarks.

\begin{table*}[t]
  \centering
  \resizebox{\textwidth}{!}{%
  \begin{tabular}{llcccccccccc}
    \toprule
    \multirow{2}{*}{\textbf{Model}} & \multirow{2}{*}{\textbf{Correction}} & \multicolumn{3}{c}{\textbf{Math}} & \multicolumn{3}{c}{\textbf{Knowledge \& Reasoning}} & \multicolumn{2}{c}{\textbf{Code}} & \textbf{Inst.} & \multirow{2}{*}{\textbf{Avg.}} \\
    \cmidrule(lr){3-5}\cmidrule(lr){6-8}\cmidrule(lr){9-10}\cmidrule(lr){11-11}
    & & MATH-500 & AIME25 & AIME26 & MMLU-Pro & GPQA-D & SuperGPQA & HumanEval+ & LCB & IFEval & \\
    \midrule
    \multirow{3}{*}{Qwen3.6-35B-A3B}
      & K2Norm & 57.73 & 21.25 & 21.67 & 68.20 & 55.62 & 43.05 & 49.31 & 19.88 & 63.13 & 44.43 \\
      & LDA & 83.72 & 58.75 & 64.16 & 73.82 & 59.22 & 49.83 & 54.73 & 22.81 & 68.40 & 59.49 \\
      & Ours & 86.72 & 55.41 & 67.92 & 75.70 & 66.54 & 52.13 & 59.45 & 26.06 & 70.28 & 62.25 \\
    \midrule
    \multirow{3}{*}{Qwen3.5-122B-A10B-FP8}
      & K2Norm & 79.97 & 11.25 & 11.67 & 72.45 & 57.20 & 51.38 & 66.69 & 25.58 & 77.06 & 50.36 \\
      & LDA & 93.40 & 51.67 & 57.50 & 80.34 & 74.69 & 60.48 & 74.31 & 36.33 & 80.75 & 67.72 \\
      & Ours & 93.72 & 47.92 & 57.09 & 80.50 & 77.90 & 61.15 & 75.46 & 36.59 & 81.42 & 67.97 \\
    \midrule
    \multirow{3}{*}{Qwen3-30B-A3B-2507}
      & K2Norm & 68.34 & 7.96 & 7.53 & 51.85 & 35.60 & 30.87 & 28.74 & 2.01 & 62.38 & 32.81 \\
      & LDA & 85.57 & 22.09 & 26.67 & 63.19 & 48.55 & 40.13 & 26.45 & 7.75 & 68.70 & 43.23 \\
      & Ours & 86.67 & 25.83 & 32.91 & 64.84 & 50.63 & 41.28 & 31.79 & 7.97 & 68.49 & 45.60 \\
    \midrule
    \multirow{3}{*}{GLM-4.7-Flash}
      & K2Norm & 17.31 & 0.42 & 0.00 & 12.81 & 15.60 & 7.39 & 10.29 & 1.46 & 30.06 & 10.59 \\
      & LDA & 70.49 & 66.67 & 64.17 & 47.46 & 33.02 & 25.46 & 59.07 & 14.00 & 59.49 & 48.87 \\
      & Ours & 74.27 & 70.83 & 63.75 & 50.89 & 32.58 & 27.86 & 62.35 & 14.22 & 62.96 & 51.08 \\
    \midrule
    \multirow{3}{*}{ERNIE-4.5-21B-A3B-PT}
      & K2Norm & 46.60 & 3.33 & 0.42 & 32.28 & 29.74 & 18.61 & 61.51 & 13.45 & 53.68 & 28.85 \\
      & LDA & 66.88 & 6.25 & 5.83 & 50.22 & 47.10 & 30.27 & 67.68 & 13.48 & 64.51 & 39.13 \\
      & Ours & 66.77 & 8.75 & 6.25 & 51.37 & 55.88 & 30.64 & 75.00 & 14.04 & 65.43 & 41.57 \\
    \bottomrule
  \end{tabular}}
  \caption{Results with ACE ($\tau$=0.7) as the skipping rule. Scores in \%.}
  \label{tab:rule-ace}
\end{table*}

\begin{table*}[t]
  \centering
  \resizebox{\textwidth}{!}{%
  \begin{tabular}{llcccccccccc}
    \toprule
    \multirow{2}{*}{\textbf{Model}} & \multirow{2}{*}{\textbf{Correction}} & \multicolumn{3}{c}{\textbf{Math}} & \multicolumn{3}{c}{\textbf{Knowledge \& Reasoning}} & \multicolumn{2}{c}{\textbf{Code}} & \textbf{Inst.} & \multirow{2}{*}{\textbf{Avg.}} \\
    \cmidrule(lr){3-5}\cmidrule(lr){6-8}\cmidrule(lr){9-10}\cmidrule(lr){11-11}
    & & MATH-500 & AIME25 & AIME26 & MMLU-Pro & GPQA-D & SuperGPQA & HumanEval+ & LCB & IFEval & \\
    \midrule
    \multirow{3}{*}{Qwen3.6-35B-A3B}
      & K2Norm & 56.48 & 19.17 & 18.33 & 67.74 & 54.04 & 42.72 & 49.09 & 20.47 & 62.56 & 43.40 \\
      & LDA & 86.08 & 64.17 & 74.58 & 75.09 & 67.49 & 51.88 & 64.40 & 26.98 & 71.49 & 64.68 \\
      & Ours & 88.10 & 63.75 & 74.58 & 75.36 & 68.50 & 53.38 & 71.57 & 26.79 & 73.03 & 66.12 \\
    \midrule
    \multirow{3}{*}{Qwen3.5-122B-A10B-FP8}
      & K2Norm & 78.31 & 14.58 & 12.50 & 71.80 & 57.52 & 51.24 & 68.37 & 24.60 & 75.67 & 50.51 \\
      & LDA & 91.48 & 58.33 & 69.58 & 79.52 & 75.63 & 59.24 & 74.54 & 34.21 & 81.78 & 69.37 \\
      & Ours & 92.85 & 58.75 & 71.67 & 79.33 & 75.88 & 59.31 & 78.66 & 35.31 & 81.62 & 70.38 \\
    \midrule
    \multirow{3}{*}{Qwen3-30B-A3B-2507}
      & K2Norm & 68.07 & 7.95 & 9.61 & 51.48 & 34.28 & 30.76 & 30.34 & 1.97 & 64.25 & 33.19 \\
      & LDA & 85.47 & 25.42 & 28.75 & 62.96 & 48.67 & 39.64 & 38.26 & 11.11 & 68.20 & 45.39 \\
      & Ours & 86.42 & 25.83 & 28.33 & 64.20 & 48.17 & 40.56 & 40.70 & 12.87 & 69.84 & 46.33 \\
    \midrule
    \multirow{3}{*}{GLM-4.7-Flash}
      & K2Norm & 13.48 & 0.00 & 0.00 & 11.74 & 15.59 & 7.25 & 6.18 & 0.62 & 29.52 & 9.38 \\
      & LDA & 67.03 & 60.00 & 57.50 & 45.18 & 31.88 & 24.27 & 55.34 & 13.30 & 57.15 & 45.74 \\
      & Ours & 70.31 & 61.25 & 62.50 & 47.81 & 30.87 & 26.59 & 56.18 & 13.16 & 60.59 & 47.69 \\
    \midrule
    \multirow{3}{*}{ERNIE-4.5-21B-A3B-PT}
      & K2Norm & 43.61 & 2.50 & 1.25 & 31.80 & 29.42 & 17.91 & 58.84 & 11.33 & 52.37 & 27.67 \\
      & LDA & 65.25 & 5.83 & 2.92 & 49.50 & 45.83 & 29.69 & 68.21 & 11.11 & 63.40 & 37.97 \\
      & Ours & 66.65 & 9.58 & 5.42 & 51.14 & 50.38 & 30.03 & 70.42 & 13.89 & 65.24 & 40.31 \\
    \bottomrule
  \end{tabular}}
  \caption{Results with Top-$p$ ($p$=0.3) as the skipping rule. Scores in \%.}
  \label{tab:rule-topp}
\end{table*}

\begin{table*}[t]
  \centering
  \resizebox{\textwidth}{!}{%
  \begin{tabular}{llcccccccccc}
    \toprule
    \multirow{2}{*}{\textbf{Model}} & \multirow{2}{*}{\textbf{Correction}} & \multicolumn{3}{c}{\textbf{Math}} & \multicolumn{3}{c}{\textbf{Knowledge \& Reasoning}} & \multicolumn{2}{c}{\textbf{Code}} & \textbf{Inst.} & \multirow{2}{*}{\textbf{Avg.}} \\
    \cmidrule(lr){3-5}\cmidrule(lr){6-8}\cmidrule(lr){9-10}\cmidrule(lr){11-11}
    & & MATH-500 & AIME25 & AIME26 & MMLU-Pro & GPQA-D & SuperGPQA & HumanEval+ & LCB & IFEval & \\
    \midrule
    \multirow{3}{*}{Qwen3.6-35B-A3B}
      & K2Norm & 81.97 & 42.08 & 41.67 & 74.27 & 61.43 & 51.27 & 70.66 & 26.75 & 71.49 & 57.95 \\
      & LDA & 89.00 & 68.33 & 75.83 & 76.20 & 67.17 & 53.58 & 74.77 & 26.61 & 73.10 & 67.18 \\
      & Ours & 93.35 & 75.41 & 79.59 & 78.23 & 70.77 & 56.93 & 78.66 & 33.41 & 76.94 & 71.48 \\
    \midrule
    \multirow{3}{*}{Qwen3.5-122B-A10B-FP8}
      & K2Norm & 93.40 & 37.50 & 43.75 & 78.39 & 68.31 & 58.59 & 82.40 & 34.06 & 81.48 & 64.21 \\
      & LDA & 96.25 & 80.00 & 85.41 & 80.42 & 77.08 & 61.65 & 85.06 & 42.14 & 86.31 & 77.15 \\
      & Ours & 96.92 & 84.17 & 88.33 & 81.22 & 78.66 & 62.71 & 86.66 & 45.80 & 86.43 & 78.99 \\
    \midrule
    \multirow{3}{*}{Qwen3-30B-A3B-2507}
      & K2Norm & 82.36 & 19.34 & 15.00 & 56.96 & 37.94 & 33.64 & 55.80 & 14.22 & 68.28 & 42.61 \\
      & LDA & 88.42 & 30.41 & 27.92 & 63.38 & 47.73 & 40.00 & 50.00 & 16.67 & 68.24 & 48.09 \\
      & Ours & 91.08 & 35.41 & 35.00 & 66.24 & 53.16 & 42.60 & 56.02 & 17.11 & 70.52 & 51.91 \\
    \midrule
    \multirow{3}{*}{GLM-4.7-Flash}
      & K2Norm & 71.00 & 53.75 & 48.75 & 43.27 & 30.99 & 22.49 & 58.46 & 12.17 & 52.17 & 43.67 \\
      & LDA & 82.03 & 82.92 & 78.75 & 59.03 & 40.59 & 34.82 & 69.59 & 20.11 & 75.80 & 60.40 \\
      & Ours & 83.72 & 86.67 & 83.75 & 60.80 & 42.74 & 35.90 & 71.87 & 18.53 & 77.45 & 62.38 \\
    \midrule
    \multirow{3}{*}{ERNIE-4.5-21B-A3B-PT}
      & K2Norm & 68.55 & 10.00 & 7.50 & 50.58 & 57.64 & 29.55 & 77.14 & 17.65 & 64.56 & 42.57 \\
      & LDA & 74.45 & 9.58 & 8.33 & 57.60 & 66.67 & 34.99 & 80.49 & 16.38 & 73.01 & 46.83 \\
      & Ours & 76.38 & 11.25 & 10.83 & 59.23 & 69.13 & 35.27 & 82.70 & 15.42 & 74.92 & 48.35 \\
    \bottomrule
  \end{tabular}}
  \caption{Results with Alloc-T ($\bar{k}$=2) as the skipping rule. Scores in \%.}
  \label{tab:rule-alloct}
\end{table*}

\begin{table*}[t]
  \centering
  \resizebox{\textwidth}{!}{%
  \begin{tabular}{llcccccccccc}
    \toprule
    \multirow{2}{*}{\textbf{Model}} & \multirow{2}{*}{\textbf{Correction}} & \multicolumn{3}{c}{\textbf{Math}} & \multicolumn{3}{c}{\textbf{Knowledge \& Reasoning}} & \multicolumn{2}{c}{\textbf{Code}} & \textbf{Inst.} & \multirow{2}{*}{\textbf{Avg.}} \\
    \cmidrule(lr){3-5}\cmidrule(lr){6-8}\cmidrule(lr){9-10}\cmidrule(lr){11-11}
    & & MATH-500 & AIME25 & AIME26 & MMLU-Pro & GPQA-D & SuperGPQA & HumanEval+ & LCB & IFEval & \\
    \midrule
    \multirow{3}{*}{Qwen3.6-35B-A3B}
      & K2Norm & 90.92 & 17.91 & 13.75 & 80.13 & 72.79 & 59.61 & 81.56 & 42.98 & 79.41 & 59.90 \\
      & LDA & 91.72 & 33.33 & 34.58 & 79.05 & 71.53 & 57.01 & 71.11 & 36.81 & 81.46 & 61.85 \\
      & Ours & 94.60 & 50.42 & 53.33 & 81.75 & 77.40 & 61.28 & 82.93 & 47.73 & 82.11 & 70.17 \\
    \midrule
    \multirow{3}{*}{Qwen3.5-122B-A10B-FP8}
      & K2Norm & 87.05 & 4.17 & 0.83 & 81.85 & 71.02 & 62.83 & 80.03 & 40.72 & 84.00 & 56.94 \\
      & LDA & 91.90 & 17.08 & 21.67 & 81.28 & 75.44 & 62.60 & 79.50 & 37.75 & 86.28 & 61.50 \\
      & Ours & 94.65 & 24.59 & 22.08 & 83.88 & 82.01 & 65.60 & 82.17 & 52.74 & 87.41 & 66.12 \\
    \midrule
    \multirow{3}{*}{Qwen3-30B-A3B-2507}
      & K2Norm & 91.57 & 20.57 & 15.47 & 70.07 & 51.96 & 47.08 & 70.35 & 25.91 & 83.12 & 52.90 \\
      & LDA & 92.20 & 20.42 & 26.67 & 70.32 & 57.39 & 45.79 & 65.78 & 27.01 & 80.48 & 54.01 \\
      & Ours & 92.65 & 24.58 & 30.83 & 72.18 & 60.17 & 50.47 & 73.17 & 37.13 & 83.38 & 58.28 \\
    \midrule
    \multirow{3}{*}{GLM-4.7-Flash}
      & K2Norm & 79.90 & 28.75 & 27.08 & 56.58 & 31.94 & 32.36 & 66.00 & 18.46 & 67.35 & 45.38 \\
      & LDA & 83.40 & 76.25 & 77.92 & 60.22 & 40.78 & 35.38 & 71.80 & 20.65 & 77.66 & 60.45 \\
      & Ours & 87.33 & 80.42 & 77.92 & 62.78 & 44.57 & 38.63 & 74.92 & 21.31 & 79.26 & 63.02 \\
    \midrule
    \multirow{3}{*}{ERNIE-4.5-21B-A3B-PT}
      & K2Norm & 75.65 & 9.17 & 5.42 & 59.43 & 68.37 & 37.32 & 80.87 & 19.22 & 72.53 & 47.55 \\
      & LDA & 74.70 & 13.33 & 10.00 & 58.20 & 63.38 & 36.45 & 79.88 & 17.51 & 74.77 & 47.58 \\
      & Ours & 80.83 & 11.25 & 10.00 & 61.89 & 72.54 & 39.41 & 82.62 & 18.79 & 77.35 & 50.52 \\
    \bottomrule
  \end{tabular}}
  \caption{Results with XShare ($\bar{k}$=2) as the skipping rule. Scores in \%.}
  \label{tab:rule-xshare}
\end{table*}

\begin{table*}[t]
  \centering
  \resizebox{\textwidth}{!}{%
  \begin{tabular}{llcccccccccc}
    \toprule
    \multirow{2}{*}{\textbf{Model}} & \multirow{2}{*}{\textbf{Correction}} & \multicolumn{3}{c}{\textbf{Math}} & \multicolumn{3}{c}{\textbf{Knowledge \& Reasoning}} & \multicolumn{2}{c}{\textbf{Code}} & \textbf{Inst.} & \multirow{2}{*}{\textbf{Avg.}} \\
    \cmidrule(lr){3-5}\cmidrule(lr){6-8}\cmidrule(lr){9-10}\cmidrule(lr){11-11}
    & & MATH-500 & AIME25 & AIME26 & MMLU-Pro & GPQA-D & SuperGPQA & HumanEval+ & LCB & IFEval & \\
    \midrule
    \multirow{3}{*}{Qwen3.6-35B-A3B}
      & K2Norm & 95.88 & 79.58 & 80.42 & 53.53 & 33.40 & 33.49 & 89.40 & 47.55 & 80.78 & 66.00 \\
      & LDA & 96.02 & 85.42 & 90.00 & 52.88 & 36.99 & 35.51 & 89.86 & 38.74 & 80.33 & 67.30 \\
      & Ours & 97.20 & 87.50 & 90.42 & 56.14 & 41.35 & 37.74 & 92.30 & 37.97 & 81.75 & 69.15 \\
    \midrule
    \multirow{3}{*}{Qwen3.5-122B-A10B-FP8}
      & K2Norm & 95.45 & 60.00 & 63.34 & 60.18 & 35.54 & 39.65 & 85.37 & 49.45 & 82.20 & 63.46 \\
      & LDA & 97.40 & 82.92 & 89.58 & 64.26 & 45.64 & 42.46 & 87.81 & 47.55 & 83.99 & 71.29 \\
      & Ours & 97.85 & 86.67 & 92.91 & 65.97 & 46.84 & 43.75 & 88.34 & 61.26 & 84.57 & 74.24 \\
    \midrule
    \multirow{3}{*}{Qwen3-30B-A3B-2507}
      & K2Norm & 90.12 & 37.08 & 33.75 & 40.31 & 28.09 & 25.55 & 86.89 & 33.48 & 77.60 & 50.32 \\
      & LDA & 92.30 & 45.42 & 48.33 & 43.87 & 31.31 & 28.12 & 84.68 & 35.01 & 78.05 & 54.12 \\
      & Ours & 94.45 & 49.58 & 55.00 & 45.55 & 32.32 & 28.70 & 88.11 & 38.45 & 78.94 & 56.79 \\
    \midrule
    \multirow{3}{*}{GLM-4.7-Flash}
      & K2Norm & 36.95 & 11.25 & 7.96 & 14.23 & 16.54 & 6.74 & 52.21 & 9.13 & 36.77 & 21.31 \\
      & LDA & 78.57 & 82.91 & 78.75 & 37.53 & 31.12 & 21.79 & 63.49 & 18.86 & 62.58 & 52.84 \\
      & Ours & 79.72 & 80.42 & 80.42 & 40.70 & 30.62 & 22.77 & 62.65 & 18.82 & 57.65 & 52.64 \\
    \midrule
    \multirow{3}{*}{ERNIE-4.5-21B-A3B-PT}
      & K2Norm & 28.73 & 2.08 & 0.00 & 18.54 & 20.52 & 11.59 & 57.93 & 12.28 & 51.32 & 22.55 \\
      & LDA & 39.05 & 1.67 & 0.42 & 28.78 & 27.52 & 14.30 & 60.37 & 12.54 & 56.40 & 26.78 \\
      & Ours & 45.32 & 1.67 & 1.25 & 29.67 & 28.03 & 18.44 & 64.10 & 11.88 & 58.22 & 28.73 \\
    \bottomrule
  \end{tabular}}
  \caption{Results with REAP ($\rho$=0.5) as the skipping rule. Scores in \%.}
  \label{tab:rule-reap}
\end{table*}

\section{Ablation Results per Skipping Rule}
\label{app:ablation}

Tables~\ref{tab:abl-rscale-rule} and~\ref{tab:abl-calib-rule} break down Tables~\ref{tab:abl-rscale} and~\ref{tab:abl-calib} by skipping rule. All results are on Qwen3.6-35B-A3B. Each cell is the mean over 8 seeds, and Avg.\ is the mean over the four benchmarks.

\begin{table*}[t]
  \centering
  \resizebox{\textwidth}{!}{%
  \begin{tabular}{lcccccccccc}
    \toprule
    \textbf{Correction} & MATH-500 & GPQA-D & HumanEval+ & IFEval & Avg. & MATH-500 & GPQA-D & HumanEval+ & IFEval & Avg. \\
    \midrule
    & \multicolumn{5}{c}{\textbf{ACE}} & \multicolumn{5}{c}{\textbf{Top-$p$}} \\
    \cmidrule(lr){2-6}\cmidrule(lr){7-11}
    LDA ($c_\ell\equiv 1$) & 83.72 & 59.22 & 54.73 & 68.40 & 66.52 & 86.08 & 67.49 & 64.40 & 71.49 & 72.37 \\
    Ours ($B$=8, $\lambda$=100) & 86.72 & 66.54 & 59.45 & 70.28 & 70.75 & 88.10 & 68.50 & 71.57 & 73.03 & 75.30 \\
    Closed form & 84.80 & 64.39 & 58.84 & 68.01 & 69.01 & 85.90 & 67.68 & 67.07 & 71.04 & 72.92 \\
    Closed form w/o calib & 85.80 & 65.15 & 60.37 & 68.98 & 70.08 & 84.40 & 61.11 & 63.41 & 69.04 & 69.49 \\
    $\lambda$=10 & 86.12 & 66.48 & 58.54 & 71.83 & 70.74 & 87.65 & 69.89 & 75.38 & 73.29 & 76.55 \\
    $\lambda$=1000 & 84.92 & 64.27 & 61.05 & 70.62 & 70.22 & 87.07 & 69.00 & 71.87 & 71.94 & 74.97 \\
    $\lambda$=$10^4$ & 85.00 & 58.34 & 56.71 & 68.61 & 67.16 & 86.03 & 66.10 & 69.51 & 70.71 & 73.09 \\
    $B$=2 & 83.92 & 65.03 & 54.96 & 71.58 & 68.87 & 87.08 & 68.43 & 71.87 & 72.49 & 74.97 \\
    $B$=4 & 85.45 & 61.68 & 53.97 & 70.26 & 67.84 & 86.12 & 67.55 & 71.57 & 73.10 & 74.59 \\
    $B$=16 & 85.32 & 67.74 & 64.25 & 71.49 & 72.20 & 87.28 & 65.53 & 70.58 & 72.14 & 73.88 \\
    \midrule
    & \multicolumn{5}{c}{\textbf{Alloc-T}} & \multicolumn{5}{c}{\textbf{XShare}} \\
    \cmidrule(lr){2-6}\cmidrule(lr){7-11}
    LDA ($c_\ell\equiv 1$) & 89.00 & 67.17 & 74.77 & 73.10 & 76.01 & 91.72 & 71.53 & 71.11 & 81.46 & 78.96 \\
    Ours ($B$=8, $\lambda$=100) & 93.35 & 70.77 & 78.66 & 76.94 & 79.93 & 94.60 & 77.40 & 82.93 & 82.11 & 84.26 \\
    Closed form & 91.80 & 70.20 & 74.69 & 76.04 & 78.18 & 93.70 & 76.77 & 82.93 & 80.64 & 83.51 \\
    Closed form w/o calib & 93.20 & 68.69 & 68.90 & 71.13 & 75.48 & 93.80 & 76.26 & 84.76 & 81.40 & 84.06 \\
    $\lambda$=10 & 93.57 & 72.60 & 79.88 & 77.29 & 80.84 & 95.20 & 77.97 & 83.31 & 82.47 & 84.74 \\
    $\lambda$=1000 & 91.93 & 68.88 & 77.21 & 75.66 & 78.42 & 94.07 & 75.82 & 83.08 & 82.35 & 83.83 \\
    $\lambda$=$10^4$ & 90.60 & 67.99 & 75.38 & 73.85 & 76.96 & 91.93 & 75.06 & 73.86 & 81.47 & 80.58 \\
    $B$=2 & 92.62 & 69.82 & 79.12 & 76.07 & 79.41 & 93.62 & 76.77 & 81.79 & 83.11 & 83.82 \\
    $B$=4 & 93.22 & 71.09 & 78.35 & 76.45 & 79.78 & 94.17 & 78.98 & 83.62 & 82.53 & 84.82 \\
    $B$=16 & 93.17 & 71.28 & 77.14 & 75.94 & 79.38 & 94.75 & 78.09 & 83.77 & 82.48 & 84.77 \\
    \midrule
    & \multicolumn{5}{c}{\textbf{REAP}} & \multicolumn{5}{c}{} \\
    \cmidrule(lr){2-6}
    LDA ($c_\ell\equiv 1$) & 96.02 & 36.99 & 89.86 & 80.33 & 75.80 &  &  &  &  &  \\
    Ours ($B$=8, $\lambda$=100) & 97.20 & 41.35 & 92.30 & 81.75 & 78.15 &  &  &  &  &  \\
    Closed form & 96.30 & 38.63 & 90.24 & 79.99 & 76.29 &  &  &  &  &  \\
    Closed form w/o calib & 96.80 & 34.85 & 87.20 & 80.03 & 74.72 &  &  &  &  &  \\
    $\lambda$=10 & 97.15 & 39.65 & 92.15 & 81.32 & 77.57 &  &  &  &  &  \\
    $\lambda$=1000 & 96.92 & 40.72 & 90.16 & 81.11 & 77.23 &  &  &  &  &  \\
    $\lambda$=$10^4$ & 96.42 & 37.81 & 89.78 & 80.23 & 76.06 &  &  &  &  &  \\
    $B$=2 & 96.50 & 38.19 & 91.92 & 81.24 & 76.96 &  &  &  &  &  \\
    $B$=4 & 97.30 & 39.77 & 91.38 & 81.81 & 77.57 &  &  &  &  &  \\
    $B$=16 & 96.97 & 41.73 & 91.76 & 81.64 & 78.03 &  &  &  &  &  \\
    \bottomrule
  \end{tabular}}
  \caption{Ablation of the scale function (Table~\ref{tab:abl-rscale}) per skipping rule on Qwen3.6-35B-A3B. Scores in \%.}
  \label{tab:abl-rscale-rule}
\end{table*}

\begin{table*}[t]
  \centering
  \resizebox{\textwidth}{!}{%
  \begin{tabular}{llcccccccccc}
    \toprule
    \multirow{2}{*}{\textbf{Rule}} & \multirow{2}{*}{\textbf{Calib.}} & \multicolumn{5}{c}{\textbf{LDA}} & \multicolumn{5}{c}{\textbf{Ours}} \\
    \cmidrule(lr){3-7}\cmidrule(lr){8-12}
    & & MATH-500 & GPQA-D & HumanEval+ & IFEval & Avg. & MATH-500 & GPQA-D & HumanEval+ & IFEval & Avg. \\
    \midrule
    \multirow{5}{*}{ACE} & C4, 2K & 82.80 & 61.49 & 52.90 & 68.97 & 66.54 & 84.42 & 67.23 & 59.07 & 70.83 & 70.39 \\
     & C4, 8K & 83.72 & 59.22 & 54.73 & 68.40 & 66.52 & 86.72 & 66.54 & 59.45 & 70.28 & 70.75 \\
     & C4, 32K & 83.75 & 61.68 & 55.95 & 69.40 & 67.69 & 85.78 & 66.29 & 59.68 & 70.45 & 70.55 \\
     & Math, 8K & 83.22 & 59.91 & 55.26 & 70.32 & 67.18 & 85.65 & 68.31 & 60.29 & 70.88 & 71.28 \\
     & Code, 8K & 84.72 & 63.95 & 63.26 & 69.12 & 70.26 & 86.25 & 69.76 & 67.15 & 70.71 & 73.47 \\
    \midrule
    \multirow{5}{*}{Top-$p$} & C4, 2K & 86.95 & 68.06 & 61.58 & 69.30 & 71.47 & 86.65 & 67.55 & 71.26 & 71.78 & 74.31 \\
     & C4, 8K & 86.08 & 67.49 & 64.40 & 71.49 & 72.37 & 88.10 & 68.50 & 71.57 & 73.03 & 75.30 \\
     & C4, 32K & 85.72 & 67.49 & 66.84 & 71.80 & 72.96 & 87.55 & 69.57 & 71.95 & 72.90 & 75.49 \\
     & Math, 8K & 86.30 & 67.74 & 67.30 & 73.40 & 73.69 & 87.03 & 67.87 & 71.65 & 73.80 & 75.08 \\
     & Code, 8K & 86.82 & 65.22 & 68.06 & 70.98 & 72.77 & 87.55 & 69.76 & 72.86 & 73.53 & 75.93 \\
    \midrule
    \multirow{5}{*}{Alloc-T} & C4, 2K & 88.67 & 67.17 & 75.31 & 71.85 & 75.75 & 92.58 & 70.71 & 78.97 & 75.09 & 79.33 \\
     & C4, 8K & 89.00 & 67.17 & 74.77 & 73.10 & 76.01 & 93.35 & 70.77 & 78.66 & 76.94 & 79.93 \\
     & C4, 32K & 89.45 & 66.48 & 77.52 & 73.25 & 76.67 & 93.58 & 71.59 & 78.89 & 76.54 & 80.15 \\
     & Math, 8K & 76.40 & 57.33 & 67.15 & 69.98 & 67.71 & 81.97 & 61.87 & 73.78 & 74.27 & 72.97 \\
     & Code, 8K & 82.90 & 60.73 & 54.57 & 69.47 & 66.92 & 87.50 & 62.37 & 65.93 & 75.30 & 72.77 \\
    \midrule
    \multirow{5}{*}{XShare} & C4, 2K & 91.35 & 71.72 & 75.61 & 80.47 & 79.79 & 94.35 & 78.22 & 83.84 & 82.63 & 84.76 \\
     & C4, 8K & 91.72 & 71.53 & 71.11 & 81.46 & 78.96 & 94.60 & 77.40 & 82.93 & 82.11 & 84.26 \\
     & C4, 32K & 92.08 & 72.28 & 74.16 & 81.12 & 79.91 & 95.15 & 76.64 & 84.91 & 82.16 & 84.72 \\
     & Math, 8K & 91.67 & 71.84 & 73.70 & 80.62 & 79.46 & 94.80 & 76.64 & 83.77 & 82.53 & 84.43 \\
     & Code, 8K & 91.42 & 69.44 & 72.71 & 78.96 & 78.14 & 94.35 & 76.33 & 84.15 & 81.85 & 84.17 \\
    \bottomrule
  \end{tabular}}
  \caption{Effect of the calibration data (Table~\ref{tab:abl-calib}) per skipping rule on Qwen3.6-35B-A3B. Scores in \%.}
  \label{tab:abl-calib-rule}
\end{table*}

\section{Skipping Budget}
\label{app:budget}

The main experiments use one budget per skipping rule (Appendix~\ref{app:rules}). Here we vary the budget on Qwen3.6-35B-A3B and evaluate on MATH-500, GPQA-Diamond, HumanEval+ and IFEval with 8 seeds. Each rule gets two budgets in addition to the main one: $\tau\in\{0.5,0.7,0.8\}$ for ACE, $p\in\{0.5,0.3,0.2\}$ for Top-$p$, $\bar{k}\in\{4,3,2\}$ for Alloc-T and XShare, and $\rho\in\{0.25,0.5,0.75\}$ for REAP, each listed from the mildest to the most aggressive budget. The main budget is the middle one for ACE, Top-$p$ and REAP and the most aggressive one ($\bar{k}=2$) for Alloc-T and XShare. For every budget, the parameters of Alloc-T, XShare and REAP are derived again with the procedures of Appendix~\ref{app:rules}, and LDA and \ours{} are calibrated again on the resulting keep masks. Table~\ref{tab:budget-q36} lists the results.

\begin{figure*}[!t]
    \centering
    \includegraphics[width=1.0\linewidth]{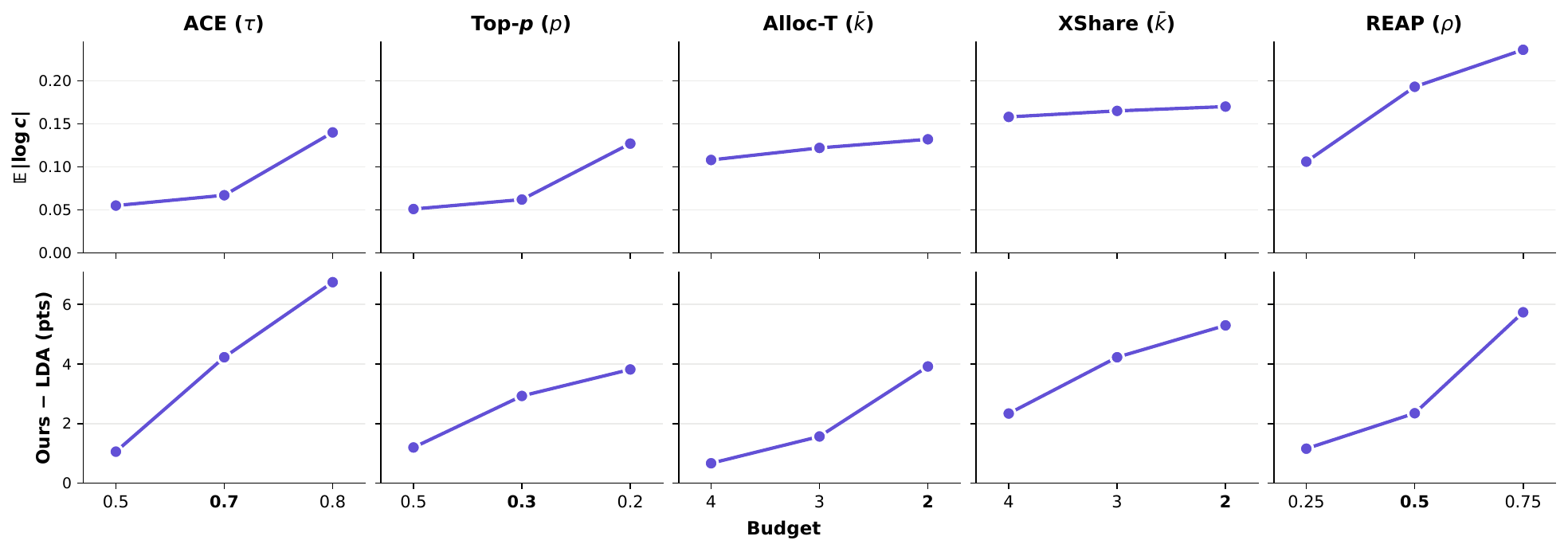}
    \caption{Skipping budget versus the departure of \ours{} from LDA on Qwen3.6-35B-A3B. Top: $\mathbb{E}|\log c|$ of the calibrated knots (Eq.~\ref{eq:app-elogc}). Bottom: gain of \ours{} over LDA in average score over MATH-500, GPQA-Diamond, HumanEval+ and IFEval. Each column is a skipping rule, the x-axis runs from the mildest to the most aggressive budget, and the bold tick is the budget of Table~\ref{tab:main}. Per-benchmark scores are in Table~\ref{tab:budget-q36}.}
    \label{fig:budget}
\end{figure*}

To measure how far \ours{} moves away from LDA, we use the mean absolute log-scale of the calibrated knots,
\begin{equation}
  \mathbb{E}|\log c| = \frac{\sum_{\ell}\sum_{b} n_{\ell,b}\,|\log c_{\ell,b}|}{\sum_{\ell}\sum_{b} n_{\ell,b}},
  \label{eq:app-elogc}
\end{equation}
where $n_{\ell,b}$ is the number of calibration tokens in bucket $b$ of layer $\ell$. LDA has $\mathbb{E}|\log c|=0$, and a value of 0.05 means that, on average over the calibration tokens, \ours{} scales the centered output of LDA by a factor of about $e^{\pm0.05}$. Figure~\ref{fig:budget} plots $\mathbb{E}|\log c|$ and the gain of \ours{} over LDA for every budget. Since the budgets of different rules are not on a common scale, we mainly compare the budgets within each rule.

\paragraph{The gain grows with the departure from $c=1$.}
Within every rule, $\mathbb{E}|\log c|$ increases from the mildest to the most aggressive budget. The gain of \ours{} moves in the same direction in all 10 budget steps. This is expected. $\mathbb{E}|\log c|$ measures how far the second moment of the LDA output departs from that of the full output. A large departure leaves more for \ours{} to correct, whereas a small one keeps the knots close to $c=1$ and leaves little room for a gain over LDA.

\paragraph{Mild and aggressive budgets.}
\ours{} improves on LDA at every budget, including the mildest one of each rule. At these budgets LDA is already within 3.09 to 5.71 points of the full model, and \ours{} gains 0.67 to 2.34 points. At the other end, at $\tau=0.8$ for ACE and $p=0.2$ for Top-$p$, LDA falls to an average of 25.06 and 25.02, with HumanEval+ below 10. \ours{} still gains 6.75 and 3.82 points, the largest gains of the two rules, but it remains far below the full model, since rescaling the pruned output cannot restore the information carried by the dropped experts.

\begin{table*}[t]
  \centering
  \footnotesize
  \setlength{\tabcolsep}{3pt}
  \begin{tabular}{llccccccc}
    \toprule
   \textbf{Rule} & \textbf{Budget} & {\boldmath$\pmb{\mathbb{E}}|\log c|$} & \textbf{Correction} & MATH-500 & GPQA-D & HumanEval+ & IFEval & \textbf{Avg.} \\
    \midrule
    \multicolumn{4}{l}{Full} & 98.10 & 83.58 & 91.31 & 86.37 & 89.84 \\
    \midrule
    \multirow{6}{*}{ACE} & \multirow{2}{*}{$\tau$=$0.5$} & \multirow{2}{*}{0.055} & LDA & 95.60 & 78.28 & 86.28 & 82.38 & 85.64 \\
     &  &  & Ours & 96.15 & 80.56 & 86.89 & 83.21 & \textbf{86.70} \\
    \cmidrule(lr){2-9}
     & \multirow{2}{*}{$\tau$=$0.7^\dagger$} & \multirow{2}{*}{0.067} & LDA & 83.72 & 59.22 & 54.73 & 68.40 & 66.52 \\
     &  &  & Ours & 86.72 & 66.54 & 59.45 & 70.28 & \textbf{70.75} \\
    \cmidrule(lr){2-9}
     & \multirow{2}{*}{$\tau$=$0.8$} & \multirow{2}{*}{0.140} & LDA & 22.76 & 23.74 & 8.31 & 45.44 & 25.06 \\
     &  &  & Ours & 35.95 & 29.61 & 11.74 & 49.96 & \textbf{31.81} \\
    \midrule
    \multirow{6}{*}{Top-$p$} & \multirow{2}{*}{$p$=$0.5$} & \multirow{2}{*}{0.051} & LDA & 96.10 & 78.28 & 86.59 & 83.50 & 86.12 \\
     &  &  & Ours & 96.52 & 80.05 & 88.19 & 84.52 & \textbf{87.32} \\
    \cmidrule(lr){2-9}
     & \multirow{2}{*}{$p$=$0.3^\dagger$} & \multirow{2}{*}{0.062} & LDA & 86.08 & 67.49 & 64.40 & 71.49 & 72.37 \\
     &  &  & Ours & 88.10 & 68.50 & 71.57 & 73.03 & \textbf{75.30} \\
    \cmidrule(lr){2-9}
     & \multirow{2}{*}{$p$=$0.2$} & \multirow{2}{*}{0.127} & LDA & 23.11 & 22.10 & 7.70 & 47.17 & 25.02 \\
     &  &  & Ours & 32.58 & 24.88 & 10.67 & 47.23 & \textbf{28.84} \\
    \midrule
    \multirow{6}{*}{Alloc-T} & \multirow{2}{*}{$\bar{k}$=$4$} & \multirow{2}{*}{0.108} & LDA & 95.89 & 78.86 & 88.76 & 83.49 & 86.75 \\
     &  &  & Ours & 96.97 & 79.36 & 89.94 & 83.42 & \textbf{87.42} \\
    \cmidrule(lr){2-9}
     & \multirow{2}{*}{$\bar{k}$=$3$} & \multirow{2}{*}{0.122} & LDA & 95.25 & 77.21 & 86.06 & 81.20 & 84.93 \\
     &  &  & Ours & 96.47 & 78.72 & 88.87 & 81.92 & \textbf{86.50} \\
    \cmidrule(lr){2-9}
     & \multirow{2}{*}{$\bar{k}$=$2^\dagger$} & \multirow{2}{*}{0.132} & LDA & 89.00 & 67.17 & 74.77 & 73.10 & 76.01 \\
     &  &  & Ours & 93.35 & 70.77 & 78.66 & 76.94 & \textbf{79.93} \\
    \midrule
    \multirow{6}{*}{XShare} & \multirow{2}{*}{$\bar{k}$=$4$} & \multirow{2}{*}{0.158} & LDA & 94.77 & 76.39 & 84.15 & 81.23 & 84.13 \\
     &  &  & Ours & 96.20 & 79.92 & 87.58 & 82.18 & \textbf{86.47} \\
    \cmidrule(lr){2-9}
     & \multirow{2}{*}{$\bar{k}$=$3$} & \multirow{2}{*}{0.165} & LDA & 93.30 & 72.66 & 78.20 & 81.48 & 81.41 \\
     &  &  & Ours & 95.32 & 78.92 & 86.21 & 82.11 & \textbf{85.64} \\
    \cmidrule(lr){2-9}
     & \multirow{2}{*}{$\bar{k}$=$2^\dagger$} & \multirow{2}{*}{0.170} & LDA & 91.72 & 71.53 & 71.11 & 81.46 & 78.96 \\
     &  &  & Ours & 94.60 & 77.40 & 82.93 & 82.11 & \textbf{84.26} \\
    \midrule
    \multirow{6}{*}{REAP} & \multirow{2}{*}{$\rho$=$0.25$} & \multirow{2}{*}{0.106} & LDA & 97.50 & 69.19 & 90.70 & 83.25 & 85.16 \\
     &  &  & Ours & 98.08 & 70.52 & 91.84 & 84.85 & \textbf{86.32} \\
    \cmidrule(lr){2-9}
     & \multirow{2}{*}{$\rho$=$0.5^\dagger$} & \multirow{2}{*}{0.193} & LDA & 96.02 & 36.99 & 89.86 & 80.33 & 75.80 \\
     &  &  & Ours & 97.20 & 41.35 & 92.30 & 81.75 & \textbf{78.15} \\
    \cmidrule(lr){2-9}
     & \multirow{2}{*}{$\rho$=$0.75$} & \multirow{2}{*}{0.236} & LDA & 51.26 & 25.63 & 78.66 & 69.27 & 56.21 \\
     &  &  & Ours & 65.62 & 25.44 & 81.02 & 75.73 & \textbf{61.95} \\
    \bottomrule
  \end{tabular}
  \caption{Skipping budgets on Qwen3.6-35B-A3B. $^\dagger$ marks the budget of Table~\ref{tab:main}. $\mathbb{E}|\log c|$ is the mean of $|\log c_{\ell,b}|$ over all knots and MoE layers, weighted by the bucket size $n_{\ell,b}$. Each cell is the mean over 8 seeds, and Avg.\ is the mean over the four benchmarks. Bold marks the correction with the higher Avg. Scores in \%.}
  \label{tab:budget-q36}
\end{table*}

\section{Inference Efficiency}
\label{app:efficiency}

We measure the serving cost of the corrections on Qwen3.6-35B-A3B with vLLM on a single NVIDIA H800 GPU. The workload consists of random prompts of 128 tokens, 1024 output tokens with EOS ignored. Each concurrency level $C$ serves $4C$ requests after $C$ warm-up requests, measured with \texttt{vllm bench serve}, and each number is the mean of two runs. We report the mean time per output token (TPOT) and the output token throughput.  We use Top-$p$ and REAP as representatives of skipping rule. The cost of the correction does not depend on the skipping rule, since $s$ and $c$ are computed from the kept gates in the same way for every rule. The budgets are those of the main experiments.

Table~\ref{tab:efficiency} shows the results. With \ours{}, the throughput speedup over the full model reaches $1.59\times$ for Top-$p$ ($C=32$) and $1.37\times$ for REAP ($C=64$). Compared with LDA, \ours{} only adds the computation of $s$ and $c$ for each token, and this overhead is negligible: under Top-$p$, where the two corrections keep almost the same number of experts, \ours{} is not slower than LDA at any concurrency, and under REAP \ours{} is at most 2\% slower than LDA. Relative to skipping without correction, \ours{} retains at least 92\% of the throughput under Top-$p$ and matches it within 1\% under REAP, so the correction preserves most of the speedup that skipping provides.

\begin{table*}[t]
  \centering
  \footnotesize
  \setlength{\tabcolsep}{4pt}
  \begin{tabular}{llcrrrrrr}
    \toprule
    & & & \multicolumn{2}{c}{$C=8$} & \multicolumn{2}{c}{$C=32$} & \multicolumn{2}{c}{$C=64$} \\
    \cmidrule(lr){4-5}\cmidrule(lr){6-7}\cmidrule(lr){8-9}
    \textbf{Rule} & \textbf{Correction} & \textbf{Kept} & TPOT$\downarrow$ & Tok/s$\uparrow$ & TPOT$\downarrow$ & Tok/s$\uparrow$ & TPOT$\downarrow$ & Tok/s$\uparrow$ \\
    \midrule
    \multicolumn{2}{l}{Full} & 8 & 9.48 & 833 & 17.07 & 1854 & 22.34 & 2835 \\
    \midrule
    \multirow{3}{*}{Top-$p$} & None & 1.90 & 6.22 & 1260 ($1.51\times$) & 10.01 & 3136 ($1.69\times$) & 13.32 & 4724 ($1.67\times$) \\
     & LDA & 2.08 & 6.48 & 1212 ($1.45\times$) & 10.91 & 2881 ($1.55\times$) & 14.69 & 4295 ($1.51\times$) \\
     & \ours{} & 2.09 & 6.28 & 1248 ($1.50\times$) & 10.68 & 2944 ($1.59\times$) & 14.43 & 4362 ($1.54\times$) \\
    \midrule
    \multirow{3}{*}{REAP} & None & 4.83 & 8.14 & 967 ($1.16\times$) & 12.84 & 2456 ($1.32\times$) & 16.33 & 3864 ($1.36\times$) \\
     & LDA & 4.66 & 8.02 & 982 ($1.18\times$) & 12.72 & 2478 ($1.34\times$) & 16.07 & 3924 ($1.38\times$) \\
     & \ours{} & 4.75 & 8.18 & 962 ($1.15\times$) & 12.72 & 2479 ($1.34\times$) & 16.27 & 3882 ($1.37\times$) \\
    \bottomrule
  \end{tabular}
  \caption{Serving efficiency of Qwen3.6-35B-A3B on one H800 GPU (128 input and 1024 output tokens) at concurrency $C$. TPOT is in ms, Tok/s is the output token throughput with the speedup over the full model in parentheses, and Kept is the mean number of routed experts per token and layer. Each number is the mean of two runs.}
  \label{tab:efficiency}
\end{table*}

\end{document}